\documentclass[10pt,twocolumn,letterpaper]{article}

\usepackage[pagenumbers]{cvpr} 

\usepackage{graphicx}
\usepackage{booktabs}
\usepackage{multirow}

\definecolor{cvprblue}{rgb}{0.21,0.49,0.74}
\usepackage[pagebackref,breaklinks,colorlinks,allcolors=cvprblue]{hyperref}
\usepackage{multirow}
\usepackage{algorithm}
\usepackage{algorithmic}
\usepackage{marvosym}

\def\paperID{11093} 
\def\confName{CVPR}
\def\confYear{2025}

\title{Scaling Down Text Encoders of Text-to-Image Diffusion Models}

\author{
    Lifu Wang\textsuperscript{\rm 1,2} \hspace{8mm} Daqing Liu\textsuperscript{\rm 1}\footnotemark[2] \hspace{8mm} Xinchen Liu\textsuperscript{\rm 1} \hspace{8mm} Xiaodong He\textsuperscript{\rm 1} \\
    \textsuperscript{\rm 1}JD Explore Academy, JD.com Inc. \hspace{4mm} \textsuperscript{\rm 2}Georgia Institute of Technology\\
    \tt\small \{lifuwang2002, daqing.liu\}@outlook.com \quad \{liuxinchen1,hexiaodong\}@jd.com\\
}

\makeatletter
\def\@fnsymbol#1{%
  \ifcase#1\or 
  \textasteriskcentered\or 
  \Letter\or 
  \textdagger\or 
  \textdaggerdbl\or 
  \textsection\or 
  \textparagraph\or 
  \textbardbl\or 
  \textasteriskcentered\textasteriskcentered\or 
  \textdagger\textdagger\or 
  \textdaggerdbl\textdaggerdbl\else\@ctrerr\fi%
}
\makeatother

\begin{document}

\maketitle

\renewcommand{\thefootnote}{\fnsymbol{footnote}}
\footnotetext[2]{Corresponding Author}

\begin{abstract}
Text encoders in diffusion models have rapidly evolved, transitioning from CLIP to T5-XXL. Although this evolution has significantly enhanced the models' ability to understand complex prompts and generate text, it also leads to a substantial increase in the number of parameters. Despite T5 series encoders being trained on the C4 natural language corpus, which includes a significant amount of non-visual data, diffusion models with T5 encoder do not respond to those non-visual prompts, indicating redundancy in representational power. Therefore, it raises an important question: ``Do we really need such a large text encoder?'' In pursuit of an answer, we employ vision-based knowledge distillation to train a series of T5 encoder models.
To fully inherit T5-XXL's capabilities, we constructed our dataset based on three criteria: image quality, semantic understanding, and text-rendering. Our results demonstrate the scaling down pattern that the distilled T5-base model can generate images of comparable quality to those produced by T5-XXL, while being 50 times smaller in size. This reduction in model size significantly lowers the GPU requirements for running state-of-the-art models such as FLUX and SD3, making high-quality text-to-image generation more accessible. Our code is available at \url{https://github.com/LifuWang-66/DistillT5}.
\end{abstract}

\section{Introduction}

\begin{figure}[t]
    \centering
    \includegraphics[width=\linewidth]{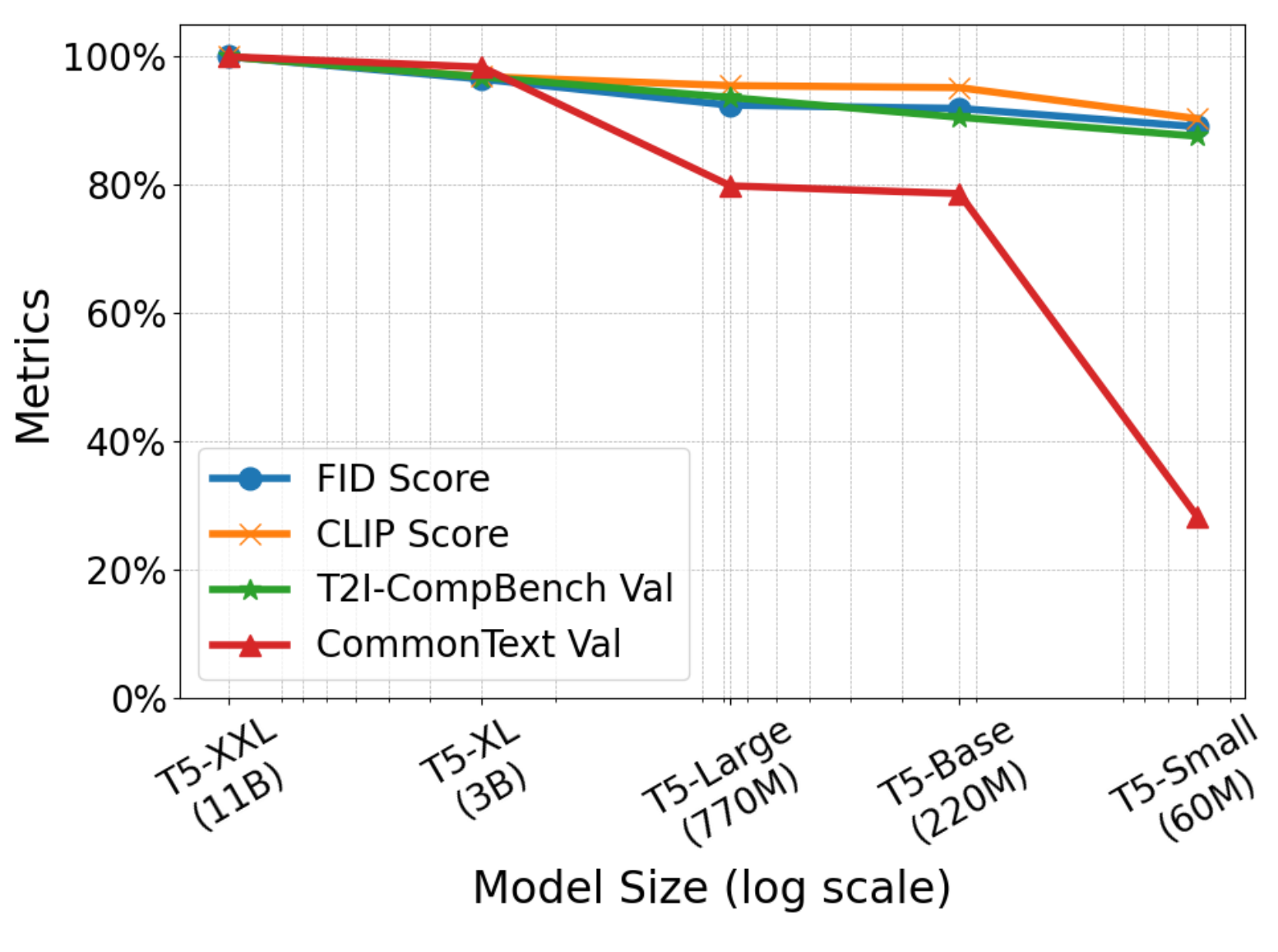}
    \caption{
        {\bf Scaling down pattern of text encoders.} We distilled T5-XXL into a series of smaller T5 models and evaluated their performance in guiding image synthesis across three key dimensions: image quality, semantic understanding, and text-rendering. Results are from section \ref{sec:scaling}. We treat T5-XXL's performance as baseline. Our findings indicate that while image quality and semantic understanding remain largely intact, text-rendering is more sensitive to reductions in model size. 
        \label{fig:law}
        }
\end{figure}

Text representation plays a crucial rule in Text-to-Image (T2I) synthesis with diffusion models \cite{sohl2015deep, song2020score,ho2020denoising,ramesh2022hierarchical,rombach2022high,saharia2022photorealistic, esser2024scaling}. Recently, text encoders used for diffusion models has a notable transition from models like CLIP \cite{radford2021learning} to the more parameter-intensive T5-XXL \cite{2020t5}. With the integration of large text encoders \cite{saharia2022photorealistic}, diffusion models are becoming increasingly sophisticated in image synthesis, especially in complex semantics comprehension and text-rendering ability. However, large models also significantly increase GPU memory requirements, presenting challenges for many users. Although previous work has quantized T5-XXL to 8-bit \cite{t5xxl8bit}, the model's parameter count remains substantial. 

\begin{figure}[t]
    \centering
    \includegraphics[width=\linewidth]{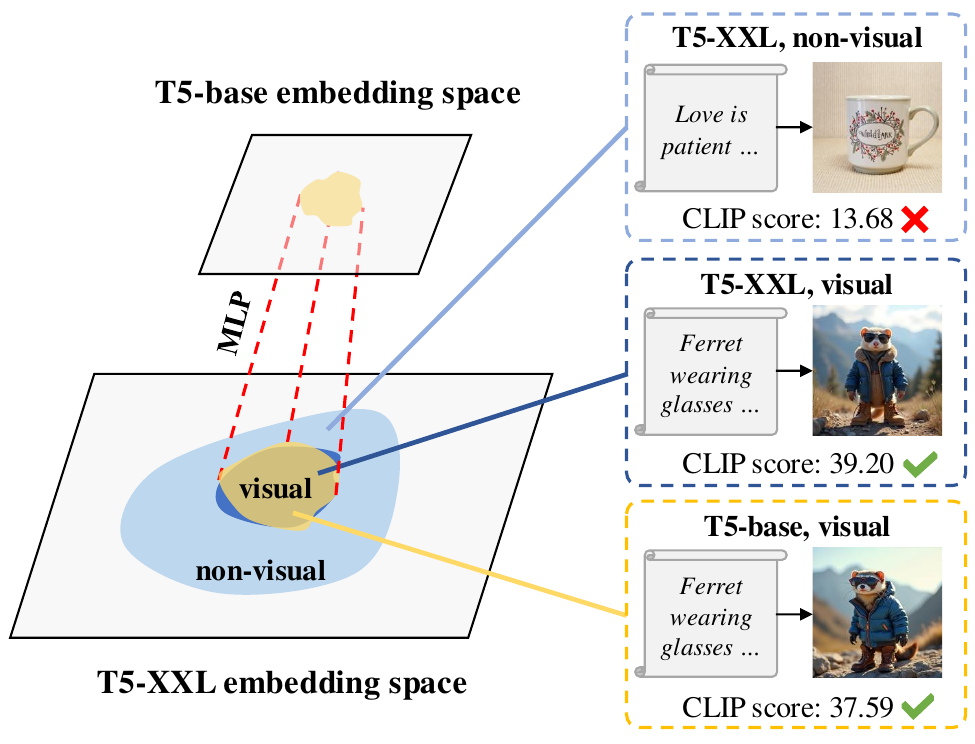}
    \caption{
        {\bf Visual and non-visual embedding space illustration.} T5 is trained on C4 dataset, in which most data are non-visual natural language. If we use a non-visual prompt to generate an image, the image does not align with the prompt well as shown by the low CLIP score. Therefore we can use a smaller model to learn the useful visual embedding and discard redundant information.
        \label{fig:intuition}
        }
    \vspace{-2mm}
\end{figure}

Our observations indicate that diffusion models using T5 encoders often struggle to generate coherent images from non-visual prompts, as illustrated in Figure \ref{fig:intuition}. We believe this is due to the T5 encoder's training on the C4 dataset \cite{2020t5}, a comprehensive natural language corpus that includes a substantial amount of non-visual data. This raises a critical question: Do we truly need such a large text encoder for effective text representation? We hypothesize that there is redundancy within the T5 encoder's embedding space particularly for T2I synthesis. If we can use a smaller text encoder to replace T5-XXL, then it means T5-XXL is overparameterized for diffusion models.

To explore the problem, we first leverage knowledge distillation \cite{hinton2015distilling, gou2021knowledge}, which proves to be an effective approach for transferring knowledge to smaller models. Traditional knowledge distillation often involves distilling logits or final outputs directly. However, we find this approach ineffective for distilling text encoders in diffusion models. Specifically, naive distillation leads to mode collapse in the student's embedding space, resulting in inaccurate prompt comprehension. Instead, we employ vision-based knowledge distillation by utilizing the impressive image synthesis capabilities of diffusion models for guidance. In addition, we discard the need of image data by using a step-following training scheme to mimic teacher's sampling trajectories. T5-XXL is widely used due to its strong ability to understand complex prompts and render text. To retain these abilities in our student model, we construct a prompt dataset based on three criteria: image quality \cite{schuhmann2022laion}, semantic understanding \cite{huang2023t2i}, and text-rendering. 

To investigate how much redundancy is in T5-XXL, we train T5 models of different size. Our results reveal that the distilled T5 model can generate images of comparable quality to those produced by T5-XXL, despite being orders of magnitude smaller in size. We find a \textbf{scaling down} pattern for text encoders, as illustrated in Figure \ref{fig:law}. The significant reduction in model size not only maintains image quality but also dramatically lowers the GPU requirements for running large diffusion models such as FLUX \cite{flux} and SD3 \cite{esser2024scaling}. Consequently, it makes high-quality text-to-image generation more accessible, opening up new possibilities to the consumer-grade GPU community. The distilled text encoders are also compatible with extra modules trained on the original diffusion model, such as ControlNet \cite{zhang2023adding} and LoRA \cite{hu2021lora}. 

In summary, the core contributions of our work are:
\begin{enumerate}
    \item We demonstrate that vision-based knowledge distillation is simple yet effective which outperforms naive distillation for distilling T5 encoders in T2I systhesis. We evaluate the distilled encoder from three perspectives: image quality, semantic understanding, and text-rendering. We show that a smaller text encoder suffice for generating high-quality images.
    \item We explore the \textbf{scaling down} pattern by training a series of T5 encoders using Flux. T5-XL retains 97\% of T5-XXL's performance while being 4 times smaller. T5-base retains T5-XXL's performance in terms of visual appeal and semantic understanding while being 50 times smaller.
\end{enumerate}

\begin{figure*}[t]
    \centering
    \includegraphics[width=1\textwidth]{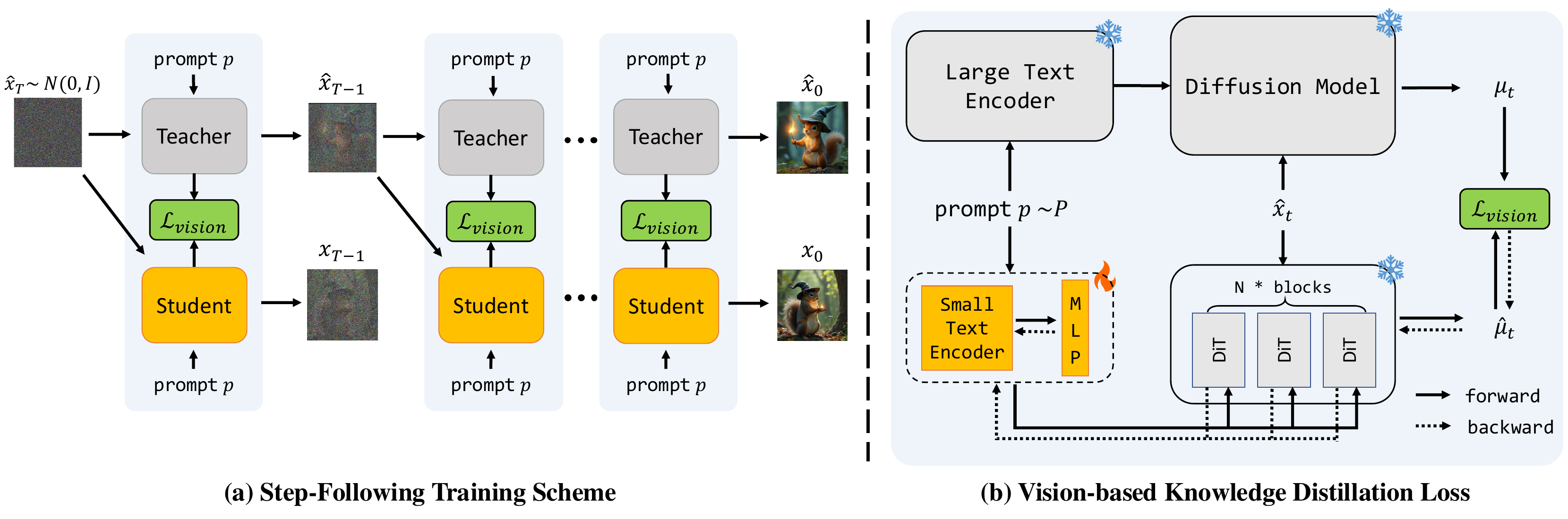}
    \caption{
        {\bf Method overview.} (a). We first illustrate the step-following distillation algorithm. Starting from a standard Gaussian noise, we pass it along with teacher embedding and student embedding to a diffusion model to obtain two predictions. We use $L_{vision}$ to train the student encoder. After obtaining the teacher latents for next timestep, we pass it to the diffusion model and repeat this process until $x_0$ is obtained. (b). During each step, we use a MLP to project student embedding to teacher's embedding space. We pass the embedding of student and teacher to a diffusion model together with $\mathbf{\hat{x}}_t$ and compute $L_{vision}$ based on predictions of the student and teacher.
        \label{fig:method}
        }
        \vspace{-3mm}
\end{figure*}

\section{Related Work}
\subsection{Text Representation in Diffusion Models}
Diffusion models \cite{song2020score,ho2020denoising} are a class of generative models that create data by iteratively refining random noise into structured outputs. In Text-to-Image synthesis \cite{esser2024scaling,rombach2022high,ramesh2022hierarchical,saharia2022photorealistic}, they work through a reverse diffusion process \cite{sohl2015deep}, where the models learn to transform noise into image under the guidance of text-embeddings through cross-attention mechanism \cite{vaswani2017attention,chen2021crossvit}. These models have attracted more attention \cite{dhariwal2021diffusion} than other generative models such as GAN \cite{goodfellow2020generative,zhu2017unpaired,kang2023scaling} and VAE \cite{kingma2013auto,razavi2019generating} due to their ability to produce high-quality and diverse images. In recent years, diffusion models have scaled a lot, from 860M \cite{rombach2022high} to 12B \cite{flux}. Text encoders used with diffusion models also have evolved from CLIP \cite{radford2021learning} to larger text encoders such as T5-XXL \cite{saharia2022photorealistic}, ChatGLM \cite{kolors}, and even large language models \cite{liu2024playground}. With larger text encoders and UNet/DiT backbones, diffusion models can generate images of good quality, but it poses challenges for deployment on consuer-grade GPUs. To address this issue, previous work focuses on reducing size of diffusion models, including pruning \cite{fang2023structuralpruningdiffusionmodels,castells2024ld}, quantization \cite{he2023efficientdm,shang2023post,li2023q}, and distillation \cite{song2024sdxs,kim2023bk}. Our work focuses on reducing the size of text encoders. We propose a simple yet effective approach that distills any text encoders into a smaller counterpart without loss of T2I synthesis performance.

\subsection{Scaling Law and Overparameterization}
Scaling law has been the norm in the field of NLP, especially in LLMs. With more parameters and training data, language models can generalize on many tasks, including language translation \cite{huang2023towards}, sentiment analysis \cite{sun2023sentiment}, question answering \cite{tan2023can}, summarization \cite{jin2024comprehensive}, and creative writing \cite{de2023writing}. The same trend was observed with text encoders used in diffusion models. Imagen \cite{saharia2022photorealistic} trained a series of UNet diffusion models with T5 of different size and found T5-XXL outperform its smaller versions. Many successors \cite{flux,esser2024scaling} followed this pattern and use T5-XXL as the to-go text encoder. However, T5 models are trained for language modeling. Prompts used to generate images only constitute a subset of the T5 training dataset Colossal Cleaned Crawled Corpus \cite{2020t5}. As such, T5-XXL might be overparameterized for encoding image prompts. Similar phenomenon have been observed in NLP, where smaller text encoders outperform LLMs on downstream tasks. LLMs have good generalizability, yet they are overparameterized for smaller tasks \cite{hsieh2023distilling,gao2023small}. We hypothesized that T5-XXL is also overparameterized for text-to-image synthesis.

\section{Method}
\subsection{Preliminary}
Diffusion models define a forward diffusion process and a reverse denoising process \cite{sohl2015deep, ho2020denoising}. The forward process gradually adds noise to the data, while the reverse process learns to denoise and recover the original data distribution.

The forward diffusion process is defined as a sequence of latent variables $\{\mathbf{x}_t\}_{t=0}^T$, starting from the original data $\mathbf{x}_0 \sim q(\mathbf{x})$. At each timestep $t$, Gaussian noise is added to the data:

\begin{equation}
q(\mathbf{x}_t \mid \mathbf{x}_{t-1}) = \mathcal{N}(\mathbf{x}_t; \sqrt{1-\beta_t} \mathbf{x}_{t-1}, (1 - \beta_t) \mathbf{I}),
\end{equation}
where $\beta_t$ is a variance schedule that controls the amount of noise added at each step. A nice property of the forward process is that we can obatin $\mathbf{x}_t$ at any time step:

\begin{equation}
\mathbf{x}_t = \sqrt{\bar{\alpha}_t} \mathbf{x}_0 + \sqrt{1 - \bar{\alpha}_t} \boldsymbol{\epsilon},
\label{eq:forward}
\end{equation}
where $\alpha_t=1-\beta_t$, $\bar{\alpha}_t = \prod_{s=1}^t \alpha_s$ and $\epsilon \sim \mathcal{N}(0, \mathbf{I})$.

The reverse process aims to recover the original data by denoising the noisy latent variables. This process is parameterized by a neural network $\mathbf{\mu}_\theta(\mathbf{x}_t, t)$ and a variance schedule $\sigma_t^2$:

\begin{equation}
p_\theta(\mathbf{x}_{t-1} \mid \mathbf{x}_t) = \mathcal{N}(\mathbf{x}_{t-1}; \mathbf{\mu}_\theta(\mathbf{x}_t, t), \sigma_t^2 \mathbf{I}),
\label{eq:backward}
\end{equation}
where $\mathbf{\mu}_\theta(\mathbf{x}_t, t)$ is the predicted mean and $\sigma_t^2$ is the variance at timestep $t$. The variance $\sigma_t^2$ can be held fixed or learned with the mean. A common choice is to use a fixed schedule.

\subsection{Vision-Based Knowledge Distillation}

Our goal is to use a smaller T5 encoder model to learn the useful visual embedding of T5-XXL. Knowledge distillation \cite{hinton2015distilling} is a commonly used technique to transfer the knowledge of a pretrained model to another smaller student model. A naive approach is to distill the output of final projection layer of T5-XXL directly:

\begin{equation}
\mathcal{L}_{naive} = \mathbb{E}_{p} \left[ \left\| \omega_\phi(p) - \omega_{\hat{\phi}}(p) \right\|^2 \right],
\end{equation}
where $\hat{\phi}$ is the teacher model, $\phi$ is the student model, $\omega$ is the output of final projection layer, and $p$ is the prompt. However, during our preliminary experiments, the naive approach often leads to mode collapse in student's embedding space, and therefore will cause misunderstanding of prompts. For more details, refer to appendix section A. In addition, it causes instability during training.

As such, we propose to use vision-based knowledge distillation by leveraging diffusion models' advanced image synthesis capability for guidance. Text-to-Image diffusion models convert prompts to its pixel/latent counterpart, which provides more fine-grained details than the T5 embedding space. Given a pretrained diffusion model $\theta$, the loss function is now:

\begin{equation}
\mathcal{L}_{vision} = \mathbb{E}_{p} \left[ \left\| \mu_\theta(\mathbf{x}_t, t, \omega_\phi(p)) - \mu_\theta(\mathbf{x}_t, t, \omega_{\hat{\phi}}(p)) \right\|^2 \right].
\label{eq:loss}
\end{equation}

\begin{algorithm}[t]
    \centering
    \caption{Step-Following Distillation}
    \label{algo:1}
    \begin{algorithmic}[1]
    \REQUIRE teacher text encoder $\omega_{\hat{\phi}}$, student text encoder $\omega_\phi$, pretrained diffusion model $\mu_\theta$, prompt dataset $\mathcal{P}$, number of inference steps $T$
    \REPEAT
        \STATE $p \sim \mathcal{P}$ 
        \STATE $\mathbf{\hat{x}}_T \sim \mathcal{N}(0, \mathbf{I})$
        \FOR{$t = T, T-1, ..., 0$}
            \STATE $\hat{\omega} = \omega_{\hat{\phi}}(p), \omega = \omega_{\phi}(p)$
            \STATE $\hat{\mu }_{t} = \mathbf{\mu}_\theta(\mathbf{\hat{x}}_{t},t,\hat{\omega})$,
                    $\mu_{t} = \mathbf{\mu}_\theta(\mathbf{\hat{x}}_{t},t,\omega)$
            \STATE $L_{\phi} = \left\| \hat{\mu}_{t} - \mu_{t}
            \right\|^2$
            \STATE $\phi \leftarrow \phi - \gamma\nabla_\phi L_\phi$
            \STATE Compute $\mathbf{\hat{x}}_{t-1}$ using Eq. (\ref{eq:backward})
        \ENDFOR
    \UNTIL {converged}
    \end{algorithmic}
\end{algorithm}

However, many state-of-the-art diffusion models \cite{flux, esser2024scaling} are trained on proprietary prompt-image datasets. It is impossible to use Eq.(\ref{eq:forward}) to obatin $\mathbf{x}_t$ for knowledge distillation. Prompts, on the other hand, are easier to acquire. Therefore, we start from $\mathbf{x}_T \sim \mathcal{N}(0, \mathbf{I})$ and use iterative denoising during training. Specifically, we pass the same latents to the pretrained diffusion model together with the student embedding and teacher embedding. After obtaining two predictions, we use \cref{eq:loss} to calculate the loss and perform back propagation. Then we obtain teacher's latent to the next time step via the scheduler and pass it to the diffusion model. We repeat the process until $\mathbf{x}_0$ is obtained, which ensures that the student model is trained on every time step. We use a Multi-Layer Perceptron to project student encoder's embedding to the teacher encoder's embedding space. We illustrate our method in Fig. \ref{fig:method} and summarize the training process in \cref{algo:1}.

\subsection{Training Data Construction}
In this section we discuss the prompt data used for knowledge distillation, which should cover all aspects of T5-XXL's visual embedding space. T5-XXL is widely adopted due to its exceptional ability to interpret complex prompts and produce coherent, contextually relevant text. To ensure that our student model inherits these capabilities, we have constructed a custom dataset based on three key criteria: image quality, semantic understanding, and text-rendering.

\noindent \textbf{Image Quality.}
For visual appeal, we use the LAION-Aesthetics-6.5+ dataset \cite{schuhmann2022laion} due to its high aesthetic score. It covers a wide range of art and photography terminologies as well as various artistic styles, allowing the model to learn nuanced stylistic details and various object-related concepts. In practice, we find using 100,000 prompts from LAION-Aesthetics-6.5+ achieves a good CLIP score \cite{radford2021learning}.

\noindent \textbf{Semantic Understanding.}
For semantic learning, we use the T2I-CompBench dataset \cite{huang2023t2i}, which includes diverse prompts covering semantic concepts like attribute-binding (color, shape, texture), spatial relationships (2D and 3D), and numeracy. We use the training set from each of the above categories, with 700 prompts per category, resulting in a total of 4,200 prompts. There are other categories such as non-spatial relationship and complex semantics, but their evaluation metrics are based on CLIP score, which overlap with our image quality metric.

\noindent \textbf{Text-Rendering.}
 We created a specialized dataset, CommonText, specifically designed to enhance the student model’s text-rendering abilities. In order to instruct our student model to generate exact text in a prompt, we generate prompts following a fixed template: \{subject\} \{action\} \{text\}, where the subject is randomly selected from one of three categories: humans, animals, or common objects. The action is also randomly chosen based on actions relevant to the subject's category. The text component includes words from the top 5,000 in the English Word Frequency dataset \cite{WordFreq}, along with 500 short sentences that covers many categories such as technology, brands, and common phrases. We generate a total of 50,000 prompts as the training set and 1,000 prompts as the validation set.
 
\begin{figure*}[!t]
    \centering
    \includegraphics[width=1\textwidth]{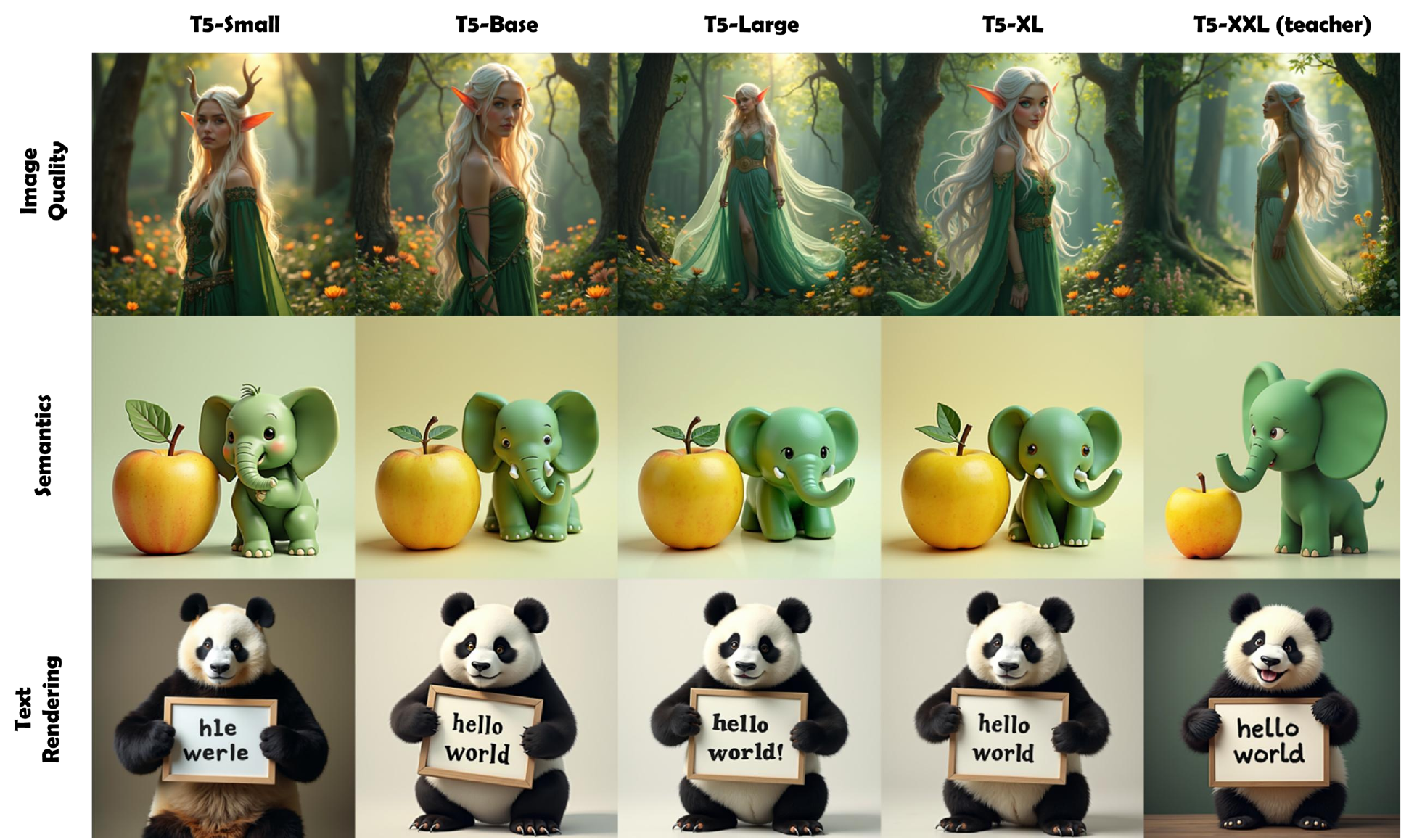}
    \caption{
        {\bf Model Size vs. Performance.} We compare images generated by T5 of different size in all three aspects. We use the same seed and guidance scale of 3.5 for inference. Text-rendering ability is affected the most by model size among three categories. Prompts: (1) ``A graceful elf ... standing in an enchanted forest under the dappled sunlight ...''; (2) ``A yellow apple and a green elephant'' (3) ``A panda presenting a board that says 'hello world'''.      
        \label{fig:comparison}
        }
        \vspace{-3mm}
\end{figure*}

\section{Experiments}
\subsection{Experimental Settings}
\label{sec:training_details}
\noindent\textbf{Training Details.}
We use Flux.1-dev as the diffusion model. Our training process consists of three stages. In the first stage, we train the student model for 50,000 iterations using the T2I-CompBench dataset, which is rich in objects and relationships, enabling the model to generate semantically meaningful images. In the second stage, we switch to the CommonText dataset for 70,000 iterations to teach the model text-rendering. In the third stage, we combine all three datasets and train the student model for 200,000 iterations. We duplicate T2I-CompBench 10 times to balance its size with the others. Training is conducted on 8 A800 GPUs with a total batch size of 32. We experiment with both 512 and 1024 resolutions for training. We find 1024 works better for text-rendering, so we use 1024 for CommonText and 512 for the others, including the combined dataset. The guidance scale is randomly sampled between 2 and 5 with interval of 0.5. We use 20 steps for iterative denoising. We use AdamW \cite{loshchilov2017decoupled} optimizer, linear learning rate scheduler, and learning rate of $1e-4$.

\noindent\textbf{Evaluation Metrics.}
We use various metrics to evaluate each criterion. To assess image quality, we test the student model's zero-shot performance on the MSCOCO-2014 validation set \cite{lin2014microsoft}, using FID \cite{heusel2017gans} and CLIP score \cite{radford2021learning} as evaluation metrics. For evaluating the semantic understanding performance, we use the T2I-CompBench validation set \cite{huang2023t2i}. To be consistent with training, we test attribute binding (color, shape, and texture), spatial relationships (2D and 3D), and numeracy. Finally, we assess the text-rendering capability of our T5-Base model using the CommonText validation set. We employ OCR to detect the text generated in images, using the text in prompts as the ground truth. Accuracy is measured at three levels of strictness: character, word, and sentence. At the character level, we align the detected sentence with the ground truth by cropping or padding it to the same length, then compare them character by character. For the word and sentence levels, a generation is considered successful if the image contains the target word or sentence.
\begin{table*}[!t]
\vspace{5pt}
\centering
\small 
\resizebox{\linewidth}{!}{
\begin{tabular} {l c c c c c c }
\toprule
\multicolumn{1}{c}{\multirow{2}{*}{\textbf{Model}}}   
  & \multicolumn{3}{c}{\textbf{Attribute Binding}}     
  & \multicolumn{2}{c}{\textbf{Spatial Relationship}} 
  & \multicolumn{1}{c}{\multirow{2}{*}{Numeracy~($\uparrow$)}} \\
              \cmidrule(lr){2-4} \cmidrule(lr){5-6}
 \multicolumn{1}{c}{}  & Color~($\uparrow$) & Shape~($\uparrow$) & Texture~($\uparrow$) & 2d-spatial~($\uparrow$) & 3d-spatial~($\uparrow$) \\
\midrule 

Pixart-Alpha &	38.91 &	40.78 &	46.18 &	20.24&	34.63&	50.29 \\
Pixart-Alpha-ft$^\dagger$ &	66.90 &	49.27 &	64.77 &	20.64&	39.01&	50.58 \\
SD 3 & 80.33&	58.31&	71.54&	31.02&	40.10 &	60.43 \\
\midrule 

Flux w/ T5-Small & 72.55 & 39.62 & 60.23 & 25.28 & 40.89 & 53.51 \\ 
Flux w/ T5-Base & 73.41 & 43.54 & 63.96 & 25.52 & 40.18 & 55.28  \\ 
Flux w/ T5-Large & 74.86 & 51.08 & 64.96 & 24.79 & 39.86 & 56.62 \\
Flux w/ T5-XL  & 77.83 & 54.86 & 66.37 & 24.94 & 40.40 & 58.43  \\

\midrule
Flux w/ T5-XXL~(teacher)  & 79.11 & 57.22 & 68.88 & 27.24 & 39.84 & 61.08  \\
\bottomrule
\end{tabular}

}
\caption{
\label{table:compbench}
{\bf Semantic understanding performance.} We use the T2I-CompBench validation set and compare our performance to other models. $^\dagger$ means the result is measured by T2I-CompBench \cite{huang2023t2i}. Our model shows comparable result compared to the teacher and other SOTA models.
}
\end{table*}
\begin{table}[t]
\centering
\small 
\resizebox{\linewidth}{!}{
\begin{tabular}{l@{\hspace{1em}}c@{\hspace{1em}}c}
\toprule
Model  & FID~($\downarrow$) & CLIP-Score~($\uparrow$) \\
\midrule
Pixart-Alpha & 28.96 & 30.61 \\
SD3 & 19.83 & 32.21 \\
\midrule
Flux w/ T5-Small & 25.10 & 28.28 \\ 
Flux w/ T5-Base & 24.32 & 29.79\\ 
Flux w/ T5-Large & 24.19 & 29.89\\
Flux w/ T5-XL & 23.17 & 30.33\\
\midrule
Flux w/ T5-XXL (teacher) & 22.36 & 31.30 \\ 
\bottomrule
\end{tabular}

}
\caption{
\label{table:clip_fid}
{\bf FID/CLIP-Score comparison on full MSCOCO-2014 validation set.} We use a guidance scale of 3.5 and resolution of 512 for Flux. We follow the default settings for other models. CLIP-Score is calculated using CLIP-ViT-g-14-laion2B-s34B-b88K.
}
\end{table}

\subsection{Scaling Down Pattern} 
\label{sec:scaling} 
To explore the scaling down pattern of text encoders specifically for image synthesis, we distill the T5-XXL model into a series of smaller T5 encoders, with sizes ranging from 60M to 11B parameters. The results of FID and CLIP are presented in Table \ref{table:clip_fid}. We also compare our results with other state-of-the-art diffusion models that use T5-XXL as text encoder. The evaluation indicates that as the size of the T5 encoder decreases, there is a decline in both FID and CLIP scores. The T5-XL configuration achieves a FID of 23.17 and a CLIP score of 30.33, outperforming the smaller T5-Small and T5-Base models. Despite the reduction in size, even the smallest model achieves a lower FID than Pixart-Alpha \cite{chen2023pixart}, suggesting that the image quality remains acceptable. As demonstrated in row 1 of Figure \ref{fig:comparison}, images generated by different T5 encoders exhibit consistent style and quality, indicating that the image quality is not significantly impacted by the reduction in model size.

\begin{table}[!t]
\centering
\small 
\resizebox{\linewidth}{!}{
\begin{tabular}{l@{\ }c@{\ }c@{\ }c}
\toprule
Model & Character~($\uparrow$) & Word~($\uparrow$) & Sentence~($\uparrow$)\\
\midrule
Pixart-Alpha & 1.7 & 1.3 & 0.0  \\
SD3 & 38.7 & 21.8 & 14.0 \\
\midrule
Flux w/ T5-Small & 31.9 & 12.0 & 9.3 \\ 
Flux w/ T5-Base & 69.3 & 42.4 & 35.6  \\ 
Flux w/ T5-Large & 69.5 & 43.8 & 36.3 \\
Flux w/ T5-XL & 77.8 & 57.7 & 48.8  \\
\midrule
Flux w/ T5-XXL~(teacher) & 76.7 & 59.3 & 51.2 \\ 
\bottomrule
\end{tabular}

}
\caption{
\label{table:ocr}
{\bf OCR detection result on our CommonText validation set}. Accuracy is measured at three levels of strictness: character, word, and sentence. Our model consistently outperforms other SOTA diffusion models.
}
\end{table}

Results on T2I-CompBench are shown in Table \ref{table:compbench}. These results indicate that across most tasks, Flux with varying T5 encoder sizes exhibit similar levels of performance exception for attribute binding. This discrepancy is likely due to the dataset size used for training. Pixart-Alpha \cite{chen2023pixart}, trained with only $1\%$ of Stable Diffusion's training data, initially shows lower performance in attribute binding. However, after fine-tuning \cite{huang2023t2i}, its attribute binding scores improve significantly, while spatial relationships and numeracy remain consistent. This suggests that increasing the size of training samples could further enhance attribute binding performance. For other semantic attributes, performance remains stable across different model capacities. As illustrated in row 2 of Figure \ref{fig:comparison}, images produced by different models demonstrate semantic consistency.

Results on text-rendering are summarized in Table \ref{table:ocr}. With almost 4 times fewer parameters, T5-Small (60M) performs considerably worse than T5-Base (220M), while T5-XL (3B) achieves scores only $1.6\%$ lower than T5-XXL (11B). It indicates that once models fall below a certain size threshold, their text-rendering capacity starts to collapse. Although T5-Base performes relatively poor compared to T5-XXL, it is much better than Pixart-Alpha and SD3. As illustrated in row 3 of Figure \ref{fig:comparison}, it can generate simple phrases, but T5-Small fails. Among all three categories, text-rendering is most significantly affected by model size. 

\subsection{Ablation Study}
\begin{figure}[t]
    \centering
    \includegraphics[width=\linewidth]{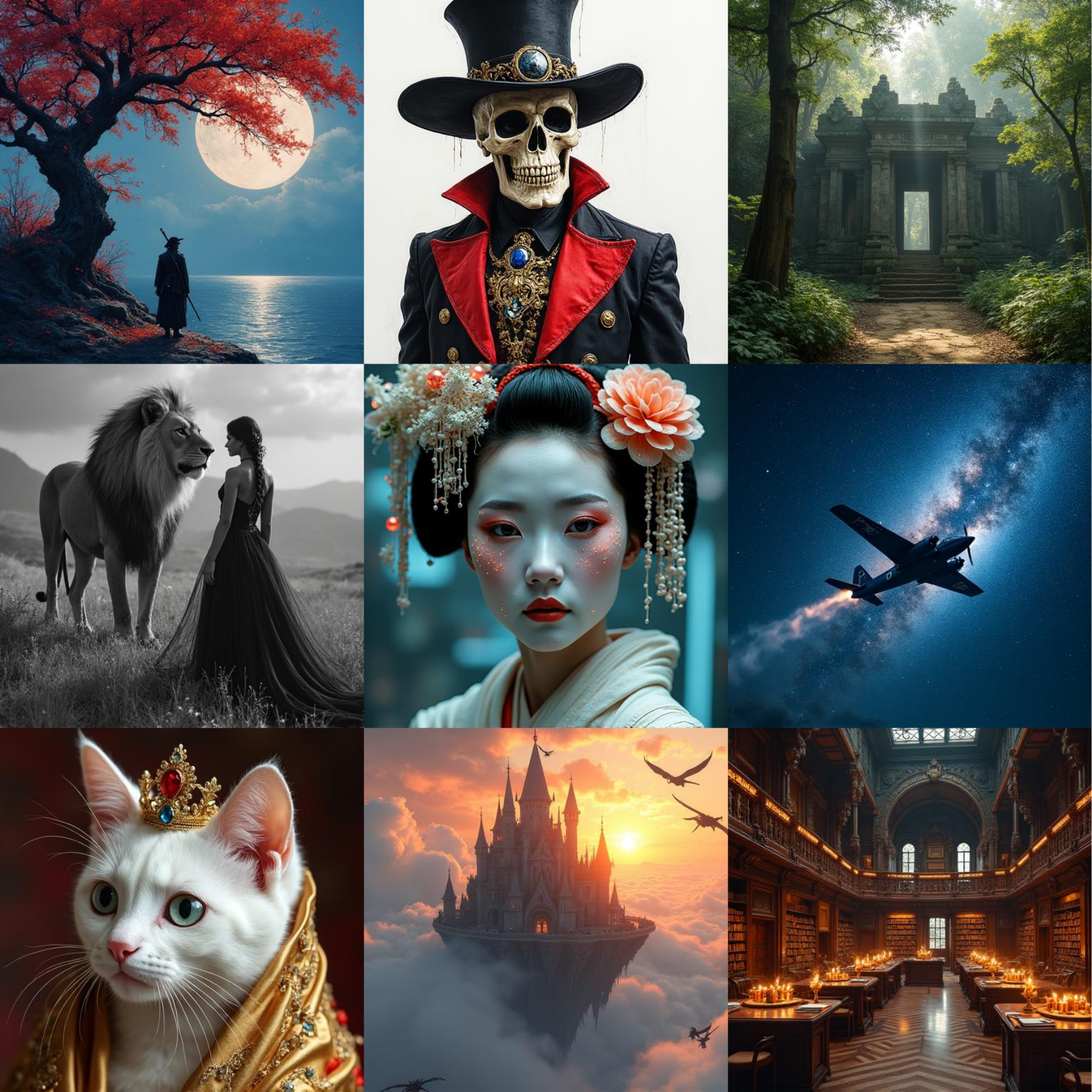}
    \caption{
         {\bf Showcase of T5-Base performance.} T5-Base can generate images with rich details and follow the prompt accurately. We put prompts in appendix for reference. 
         }
    \label{fig:showcase}    
\end{figure}

We analyze the impact of different training datasets. While T5-XL retains much of T5-XXL’s capability, it remains large. Hence, we select T5-Base for a balance of size and performance. As shown in Figure \ref{fig:showcase}, it generates highly detailed images across various concepts and styles. We evaluate T5-Base on the three categories in Section \ref{sec:training_details}, with results in Table \ref{table:ablation}. Training on LAION-Aesthetics-6.5+ yields a high CLIP score but weakens semantic understanding and text-rendering. T2I-CompBench improves semantics but still struggles with text. CommonText enhances text-rendering but reduces visual appeal and semantics. Combining all three datasets achieves the best overall performance. We also tested a model trained with naive distillation, which performed significantly worse than our vision-based distillation.

\begin{table*}[t]
\centering
\small 
\resizebox{\linewidth}{!}{
\begin{tabular} {c | c c c | c c c c}
\toprule
 $\mathcal{L}_{vision}$ & Image Quality & Sem. Und. & Text Rendering & FID~($\downarrow$) & CLIP-Score~($\uparrow$) & T2I-Compbench Val($\uparrow$) & CommonText Val~($\uparrow$) \\
\midrule 
    \checkmark & \checkmark  &  &  & 24.13 & 29.69  & 31.09 & 2.97 \\
    \checkmark &  & \checkmark  &  & 23.55 & 27.88  & 44.93 & 1.32 \\
    \checkmark &  &  & \checkmark  & 28.95 &  25.62 & 21.20 & 43.41 \\
\midrule 
     & \checkmark & \checkmark & \checkmark & 26.47 & 22.52  & 13.78 & 0.35 \\ 
    \checkmark & \checkmark & \checkmark & \checkmark & 24.32 & 29.79 & 50.32 & 49.1 \\ 
\bottomrule
\end{tabular}
}
\caption{
\label{table:ablation}
{\bf Ablation Study.} We test the effect of different components of our training dataset. We use the average score for T2I-CompBench and CommonText validation set. We also evaluate our vision distill loss against the naive embedding distill loss.
}
\end{table*}


\subsection{Embedding Space Analysis}
\begin{figure}[t]
    \centering
    \begin{subfigure}[b]{0.23\textwidth}
        \centering
        \includegraphics[width=\textwidth]{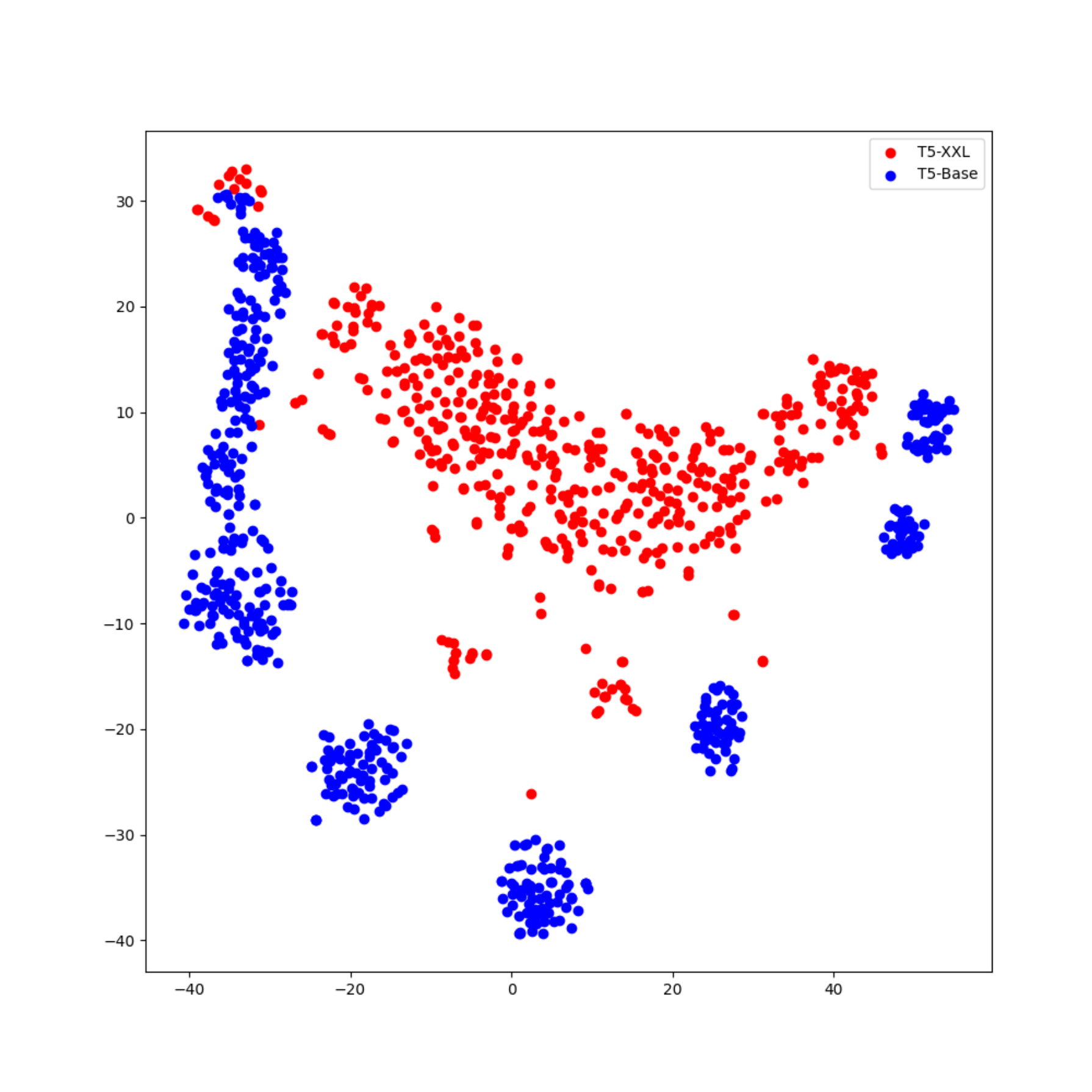}
         \vspace{-5mm}
        \caption{}
    \end{subfigure}
    \begin{subfigure}[b]{0.23\textwidth}
        \centering
        \includegraphics[width=\textwidth]{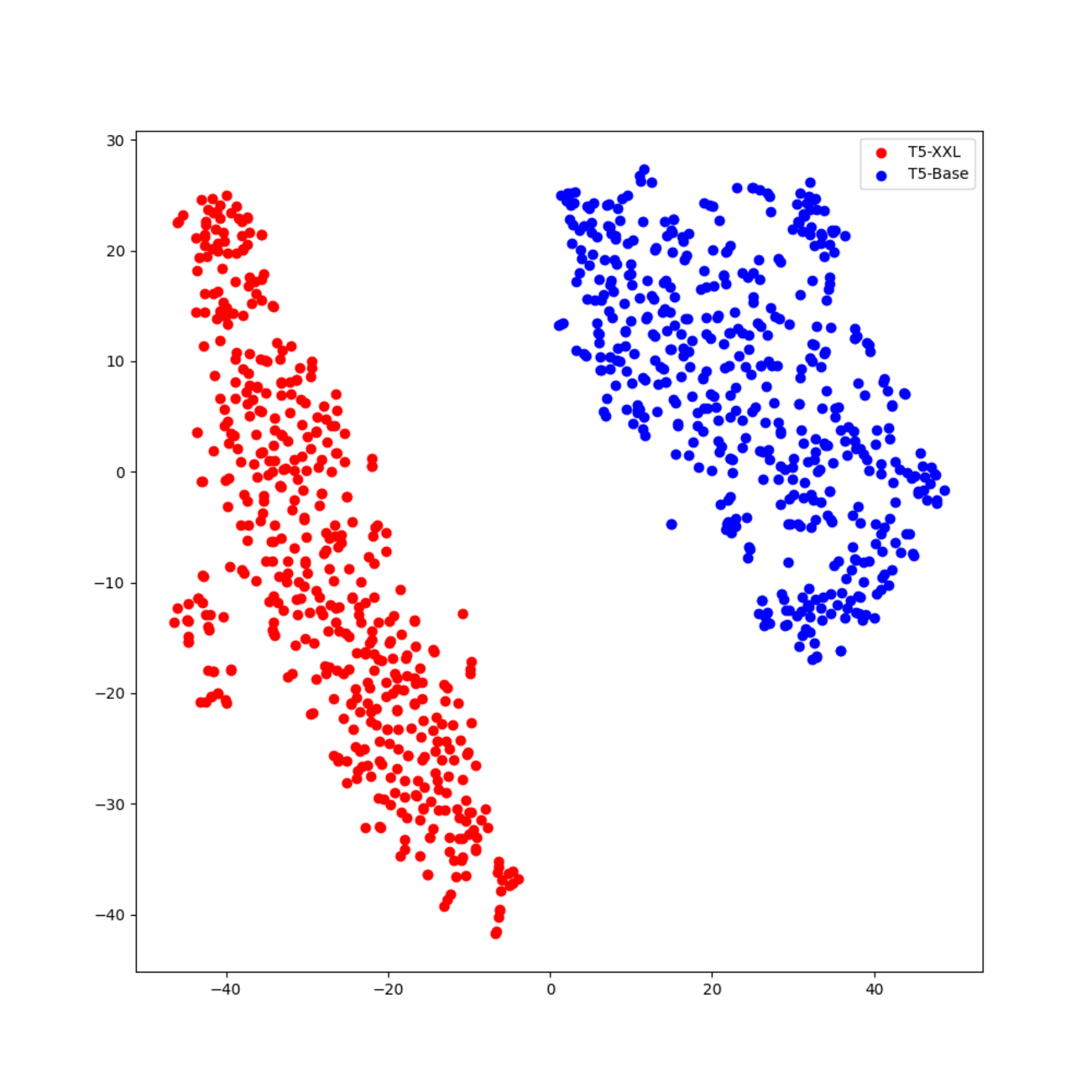}
        \vspace{-5mm}
        \caption{}
    \end{subfigure}
    \caption{
        t-SNE visualization of \textcolor{blue}{T5-Base} and \textcolor{red}{T5-XXL} embeddings. (a) and (b) shows naive distillation and vision-based distillation respectively. Naive distillation suffers from mode collapse, while vision-based distillation covers a wider range of embeddings.
    }
    \label{fig:tsne}
\end{figure}

\begin{figure}[t]
    \centering
    \includegraphics[width=\linewidth]{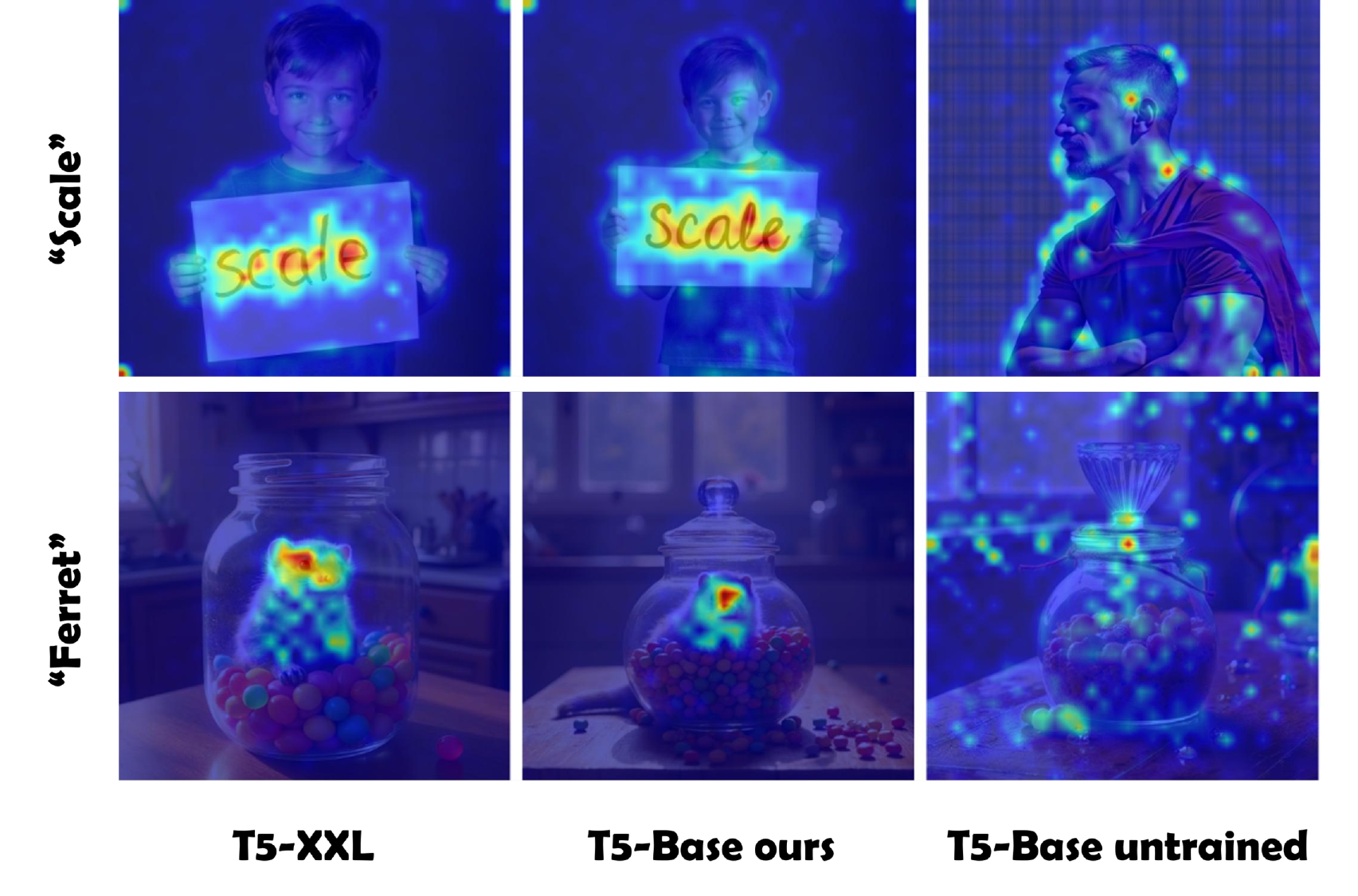}
    \caption{
        {\bf Visialization of cross-attention map.} We visualize the cross-attention map of the token `scale' (row 1) and `ferret' (row 2). (a): T5-XXL; (b): T5-Base ours; (c) T5-Base w/o training. The maps are overlaid on the corresponding images. Our T5-Base successfully learns to guide image generation, though it has a different distribution from its teacher.
        \label{fig:attention_map}
        }
\end{figure}

Our results show that T5-Base has learned most T2I capabilities of T5-XXL, but does T5-Base learn the distribution of T5-XXL? To investigate this, we randomly sampled 500 prompts from DiffusionDB \cite{wang2022diffusiondb}, a comprehensive text-to-image dataset, and processed them using both T5-XXL and our T5-Base model. We then reduced their output embeddings using t-SNE \cite{van2008visualizing}  and visualized the distributions in Figure \ref{fig:tsne}. It is evident that there is a significant gap between their distributions, suggesting that T5-Base does not replicate the exact distribution of T5-XXL. This observation highlights why naive distillation fails when distilling T5 for image synthesis. Due to the substantial size difference, T5-Base cannot establish a one-to-one mapping between its embedding space and that of T5-XXL. Naive distillation enforces such an exact mapping, causing T5-Base to predict some common embeddings. Conversely, vision-based distillation introduces uncertainty during training through noise and more detailed latent space features. Although its embedding space no longer overlaps with T5-XXL's, it offers meaningful guidance to the diffusion model via cross-attention, as illustrated in Figure {\ref{fig:attention_map}}.

\subsection{Compatibility with Auxiliary Modules}
\begin{figure}[t]
  \centering
  \begin{subfigure}[b]{\linewidth}
    \includegraphics[width=\linewidth]{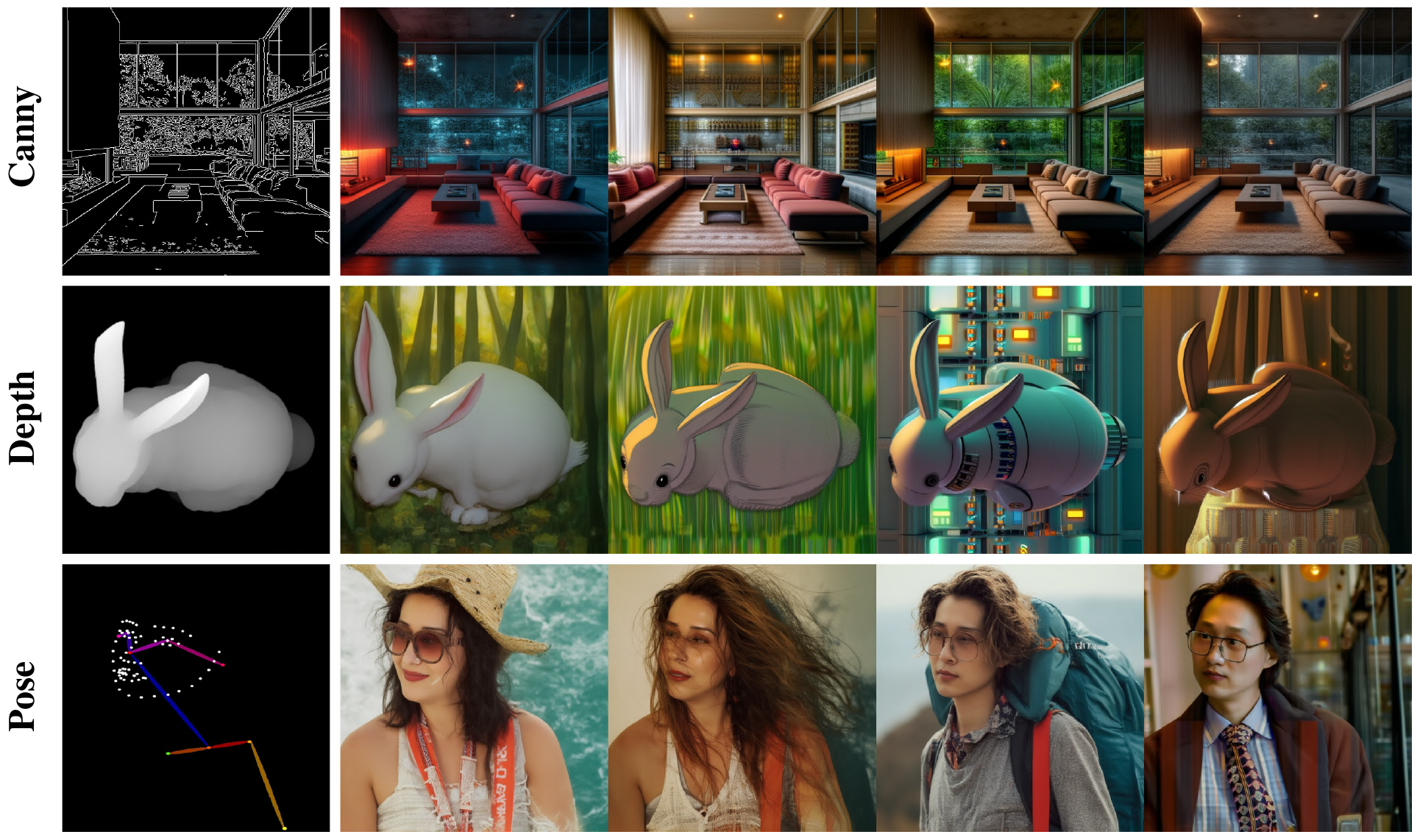}
    \caption{\textbf{Compatibility with ControlNet}}
    \label{fig:controlnet}
  \end{subfigure}
  \hfill
  \begin{subfigure}[b]{\linewidth}
    \includegraphics[width=\linewidth,page=1]{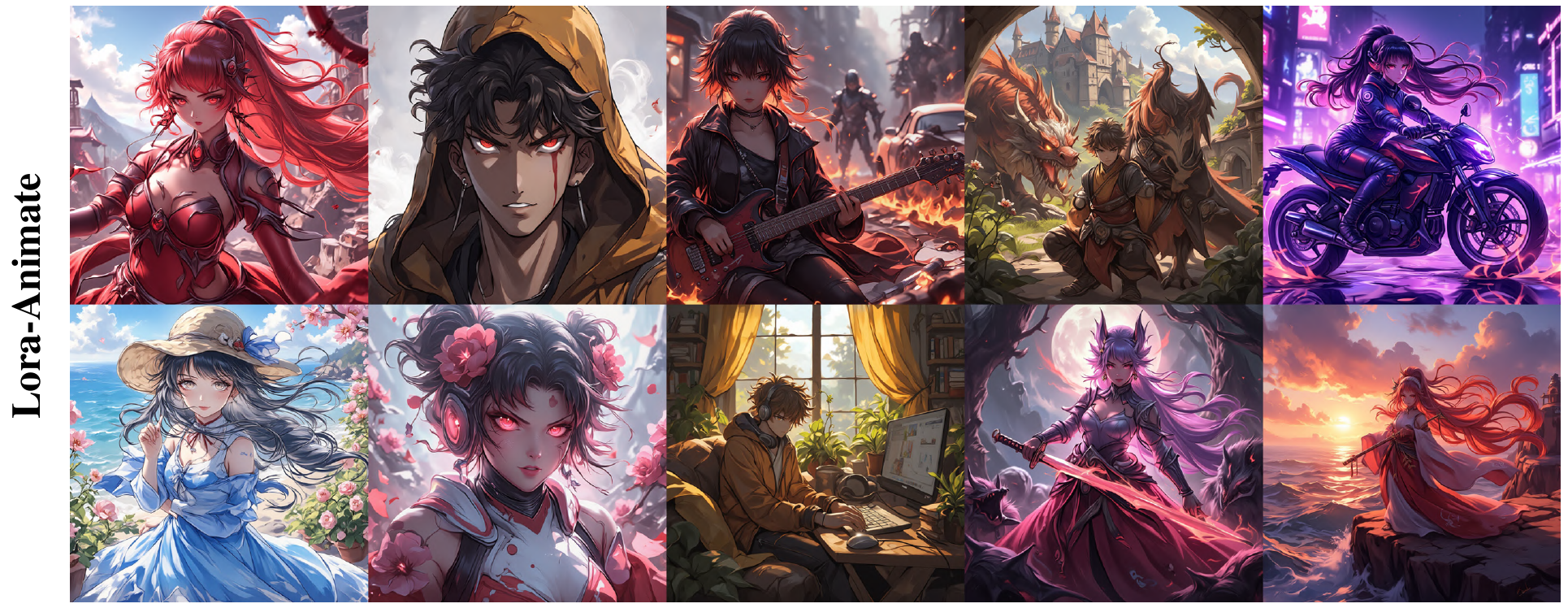}
    \caption{\textbf{Compatibility with LoRA}}
    \label{fig:lora}
  \end{subfigure}
  \begin{subfigure}[b]{\linewidth}
    \includegraphics[width=\linewidth,page=2]{figures/lora_schnell.pdf}
    \caption{\textbf{Compatibility with Distilled Model}}
    \label{fig:schnell}
  \end{subfigure}
  \caption{
  \textbf{Compatibility with Auxiliary Modules.}
  (a) We test ControlNet with Canny edge (row 1), depth map (row 2), and OpenPose (row 3).
  (b) We test LoRA with anime-style prompts.
  (c) We test step-distilled diffusion model by comparing results using T5-XXL (top) and T5-Base (bottom). We use guidance scale of 0 and 4 inference steps.
  }
  \label{fig:compatibility}
\end{figure}

Beyond image synthesis using the base diffusion model, we evaluate the compatibility of our T5-Base with other auxiliary modules. First, we integrate our T5-Base with ControlNet \cite{zhang2023adding}, which enhances image generation with additional conditions.  We use Shakker-Labs/FLUX.1-dev-ControlNet-Union-Pro \cite{fluxcontrolnet} and test it with Canny edge, depth map, and OpenPose conditions. Sample images are presented in Figure \ref{fig:controlnet}, demonstrating that our model can provide diverse style guidance alongside these image conditions. Next, we assessed its compatibility with LoRA \cite{hu2021lora}, a method that efficiently updates the weights of the base diffusion model. We use prithivMLmods/Canopus-LoRA-Flux-Anime \cite{fluxlora}. Results depicted in Figure \ref{fig:lora} indicate that our text encoder effectively guides a diffusion model with fine-tuned weights. Finally, we examined whether T5-Base is compatible with a step-distilled model. We use Flux.1-Schnell \cite{fluxschnell} since it was trained from our based diffusion model. As shown in Figure \ref{fig:schnell}, the comparison with T5-XXL indicates that T5-Base is also capable of guiding step-distilled models effectively.

\begin{table}[t]
\centering
\small 
\resizebox{\linewidth}{!}{
\begin{tabular}{l@{\ }c@{\ }c@{\ }c@{\ }c}
\toprule
\multicolumn{1}{c}{\multirow{2}{*}{\textbf{Model}}}   
  & \multicolumn{2}{c}{\textbf{Pipeline Latency}}     
  & \multicolumn{2}{c}{\textbf{Memory}} \\
              \cmidrule(lr){2-3} \cmidrule(lr){4-5}
 \multicolumn{1}{c}{}  & w/o offload & w/ offload & T5 & Pipeline\\
\midrule 
T5-Base & 8.89s & 16.61s & 285MB & 23401MB \\ 
T5-XXL & 8.95s & 24.71s & 11004MB & 34113MB \\ 
\bottomrule
\end{tabular}
}
\caption{
\label{table:performance}
{\bf Memory and time efficiency improvement.} We measure the latency using Flux pipeline with 8 steps on an A100 GPU. The performance would be boosted if we do not need CPU offload.
}
\end{table}

\subsection{Memory and Time Efficiency}
Diffusion models are becoming increasingly large. The Flux pipeline \cite{flux} itself exceeds 24GB, let alone extra space to store activations during inference. It makes running the pipeline on consumer-grade GPUs difficult. By replacing T5-XXL with our T5-Base, we can reduce the size of Flux pipeline to 24GB. Combining with other strategies such as quantization \cite{he2023efficientdm,shang2023post,li2023q} and pruning \cite{fang2023structuralpruningdiffusionmodels,castells2024ld} of diffusion models, we can run Flux on consumer-grade GPUs such as RTX 4090 without using CPU offload. Table \ref{table:performance} shows the latency of running Flux pipeline with T5-Base and T5-XXL. Without CPU offload, T5-Base has over 2.7 times speed up compared to T5-XXL with CPU offload.

\section{Conclusion}
In this paper, we explored the critical question of whether large text encoders like T5-XXL are necessary for effective text-to-image synthesis in diffusion models. By utilizing a novel vision-based knowledge distillation approach, we successfully demonstrate that smaller text encoders can achieve comparable performance to T5-XXL while significantly reducing model size and computational requirements, making it a viable and efficient alternative for state-of-the-art diffusion models. More importantly, we investigated the scaling down pattern for text encoders and discovered that text-rendering is the most affected attribute when the model size changes. Our distilled text encoders are also compatible with additional modules like ControlNet and LoRA, ensuring broad applicability and integration with existing diffusion models.
Overall, our work paves the way for more accessible T2I synthesis by integrating text encoders into the realm of efficient diffusion models. We hope that this aspect will receive greater attention from the community in future research.

{
\small
\bibliographystyle{ieeenat_fullname}
\bibliography{main}
}

\clearpage
\setcounter{figure}{8}
\setcounter{table}{5}
\appendix 

\noindent In this supplementary material, we present more details, experiments, qualitative results and disccusions that not covered in the main text.
\begin{itemize}
\item \cref{sup:details} provides \textbf{more details} for \textit{mode collapse}, \textit{qualitative evaluation}, \textit{model architecture}, \textit{training data} (\cref{fig:data}), and \textit{figure details}.
\item \cref{sup:pixart} describes \textbf{more experiments}  conducted on PixArt-Alpha (\cref{table:pixart_compbench,table:pixart_clip_fid}, \cref{fig:pixart_comparison}) to assess the \textit{generalizability} of our method.
\item \cref{sup:qualitative} provides \textbf{more qualitative results} on \textit{image quality} (\cref{fig:image_quality_comparison}), \textit{semantic understanding} (\cref{fig:semantic_comparison}), and \textit{text rendering} (\cref{fig:text_rendering_comparison}) comparison across different T5 size.
\item \cref{sup:limitation} discusses our \textbf{limitations \& social impact}.
\item \cref{sup:prompts} lists the \textbf{prompts used} to generate the images featured in the main text.
\end{itemize}

\section{More Details}
\label{sup:details}

\begin{figure}[!t]
    \centering
    \begin{subfigure}[b]{0.15\textwidth}
        \centering
        \includegraphics[width=\textwidth]{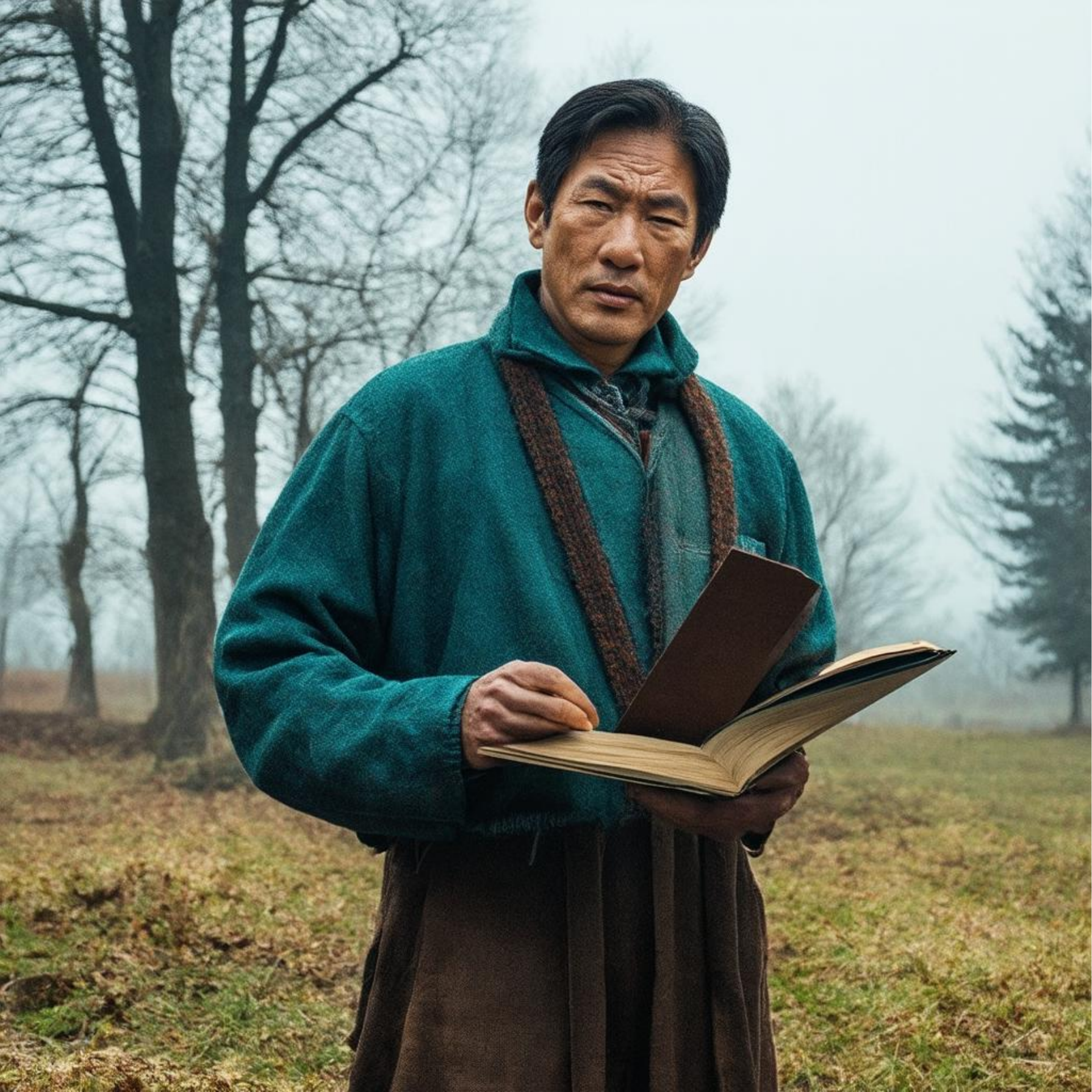}
         \vspace{-5mm}
        \caption{}
    \end{subfigure}
    \begin{subfigure}[b]{0.15\textwidth}
        \centering
        \includegraphics[width=\textwidth]{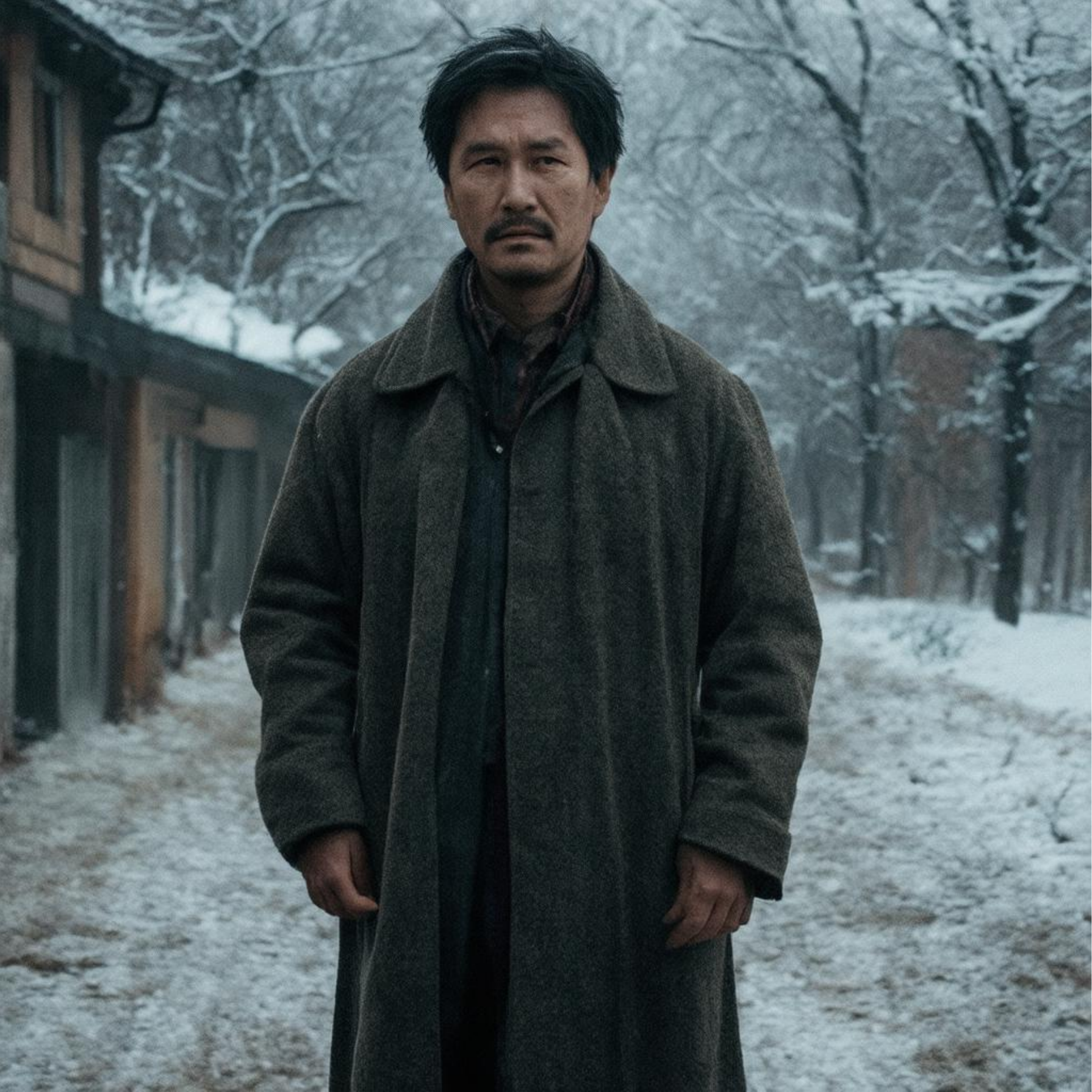}
        \vspace{-5mm}
        \caption{}
    \end{subfigure}
    \begin{subfigure}[b]{0.15\textwidth}
        \centering
        \includegraphics[width=\textwidth]{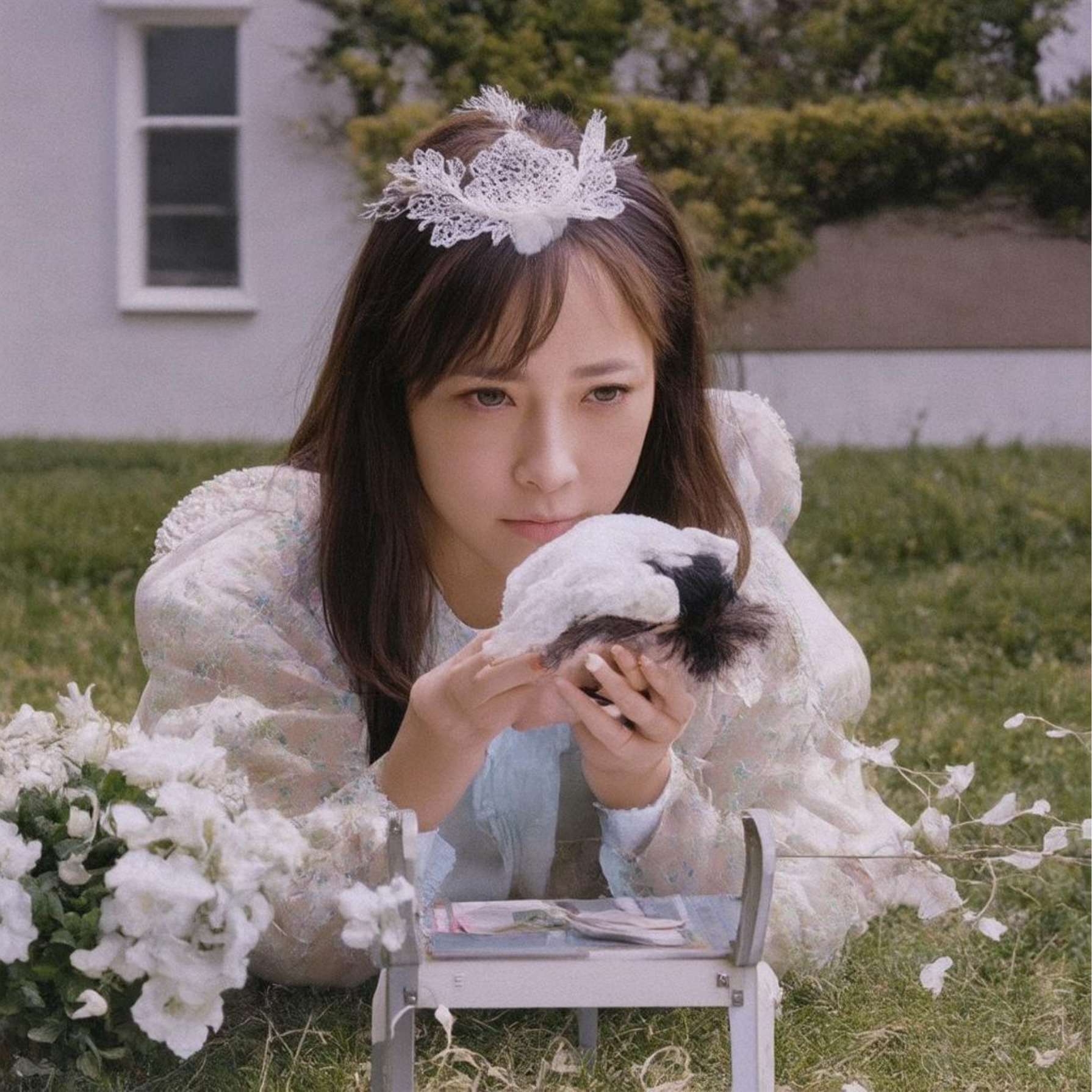}
        \vspace{-5mm}
        \caption{}
    \end{subfigure}
    \vspace{-2mm}
    \caption{
        {\bf Mode collapse caused by naive distillation.} The images were generated using the prompts (a) `A rat,' (b) `A cat,' and (c) `A monkey.' Due to mode collapse, T5 represents `rat,' `cat,' and `man' with the same embedding, and `monkey' and `woman' with the same embedding.
    }
    \label{fig:mode_collapse}
    \vspace{-5mm}
\end{figure}

\noindent\textbf{Visualization of Mode Collapse.} When text embeddings collapse into several modes, the text encoder will represent two different concepts, such as rat and man, with the same embedding. As illustrated in Fig. \ref{fig:mode_collapse}, naïve distillation makes T5 predict the wrong embedding and therefore the DM generates a failed image.

\noindent\textbf{Qualitative evaluation.} We conducted a user study with 20 participants, presenting them with 15 two-image-text pairs for comparison. Participants were asked to evaluate which pair aligned better with the prompt: A, B, or if they were tied. The results indicated that 82.7\% believed the pairs were tied, 12.0\% favored T5-XXL, and 5.3\% favored T5-Base.

\noindent\textbf{Model Architecture.} To project the student encoder's embedding to the T5-XXL's embedding space, we use a Multi-Layer Perceptron (MLP). The MLP consists of a Linear layer with input dimension of student's embedding dimension ($512$ for T5-Small , $768$ for T5-Base, $1024$ for T5-Large, and $2048$ for T5-XL) and output dimension of $4096$, followed by a ReLU \cite{nair2010rectified} activation layer, a dropout layer \cite{srivastava2014dropout} with dropout rate of $0.1$, and a final Linear layer with input dimension $4096$ and output dimension $4096$.

\begin{figure}[!t]
    \centering
    \includegraphics[width=\linewidth]{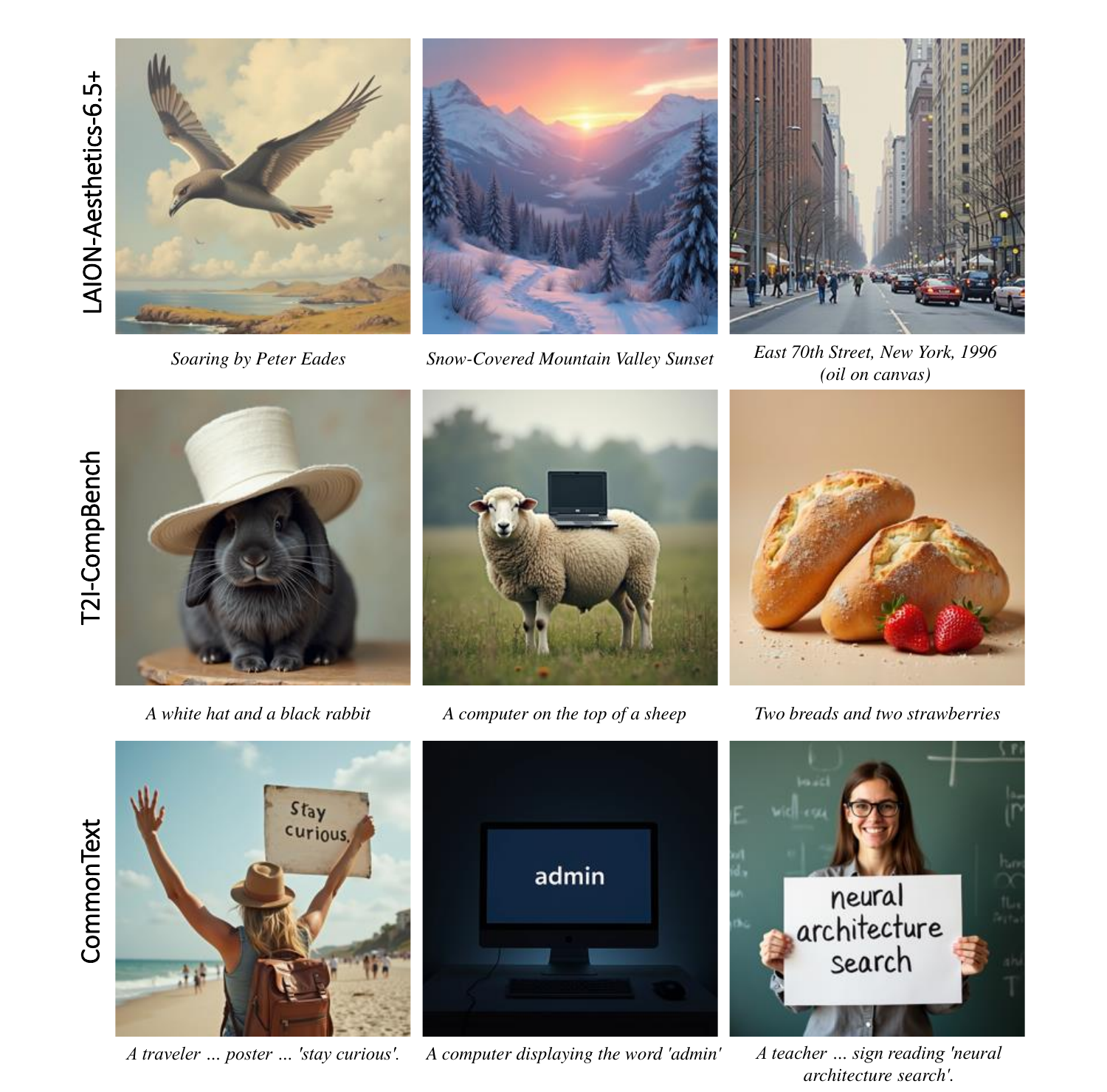}
    \caption{
        {\bf Training data samples.} Prompts are sampled from each dataset and images are generated using Flux.
        \label{fig:data}
        }
\end{figure}

\noindent\textbf{Dataset Samples.} We visualize the three components of training prompt data in Figure \ref{fig:data}. Each component highlights a specific ability of the T5-XXL model, providing a comprehensive guide for our student model to learn diverse and complex embeddings.

\noindent\textbf{Figure Details.} In Figure 1, the y-axis represents the score of student models as a percentage of the T5-XXL model's score, which is set as $100\%$. For score of T2I-CompBench and CommonText validation set, we take the average across all categories. The x-axis displays logarithm of the number of parameters in each student model.

\noindent For Figure 6, we randomly sample 500 prompts from DiffusionDB \cite{wang2022diffusiondb} and pass them to our T5-Base and T5-XXL respectively. we compute the mean of the embeddings across the sequence dimension, resulting in 500 data points for each model, each with 4096 dimensions. To facilitate visualization, we apply t-SNE \cite{van2008visualizing} to reduce these high-dimensional data points to 2 dimensions. The reduced data are then plotted on a 2D plane, with the t-SNE components on the x and y axes, allowing us to compare how the T5-Base and T5-XXL models represent the prompts in a lower-dimensional space. 

\noindent For Figure 7, we extract the attention proability of the attention layers of the $10^{th}$ inference step. Flux utilizes the same architecture as SD3 \cite{esser2024scaling}, which combines self-attention and cross-attention into a unified large attention matrix. As such, the matrix is of shape $[B,S,H+E,H+E]$, where $B$ is batch size, $S$ is sequence length, $H$ is hidden states dimension, and $E$ is encoder hidden states dimension. We extract the upper right corner ($[B,S,:E,E:]$) of the attention map, which corresponds to cross-attention in previous UNet/DiT \cite{rombach2022high,chen2023pixart} structured diffusion models. For the visualization of a specific token, we extract the attention map for that token by indexing the sequence dimension. We then upsample the attention map from the latent space to the image space and overlay it onto the generated image, providing a clear view of how attention is distributed for that token in the final output.

\section{More Experiments}
\label{sup:pixart}

\begin{figure*}[!t]
    \centering
    \includegraphics[width=1\textwidth]{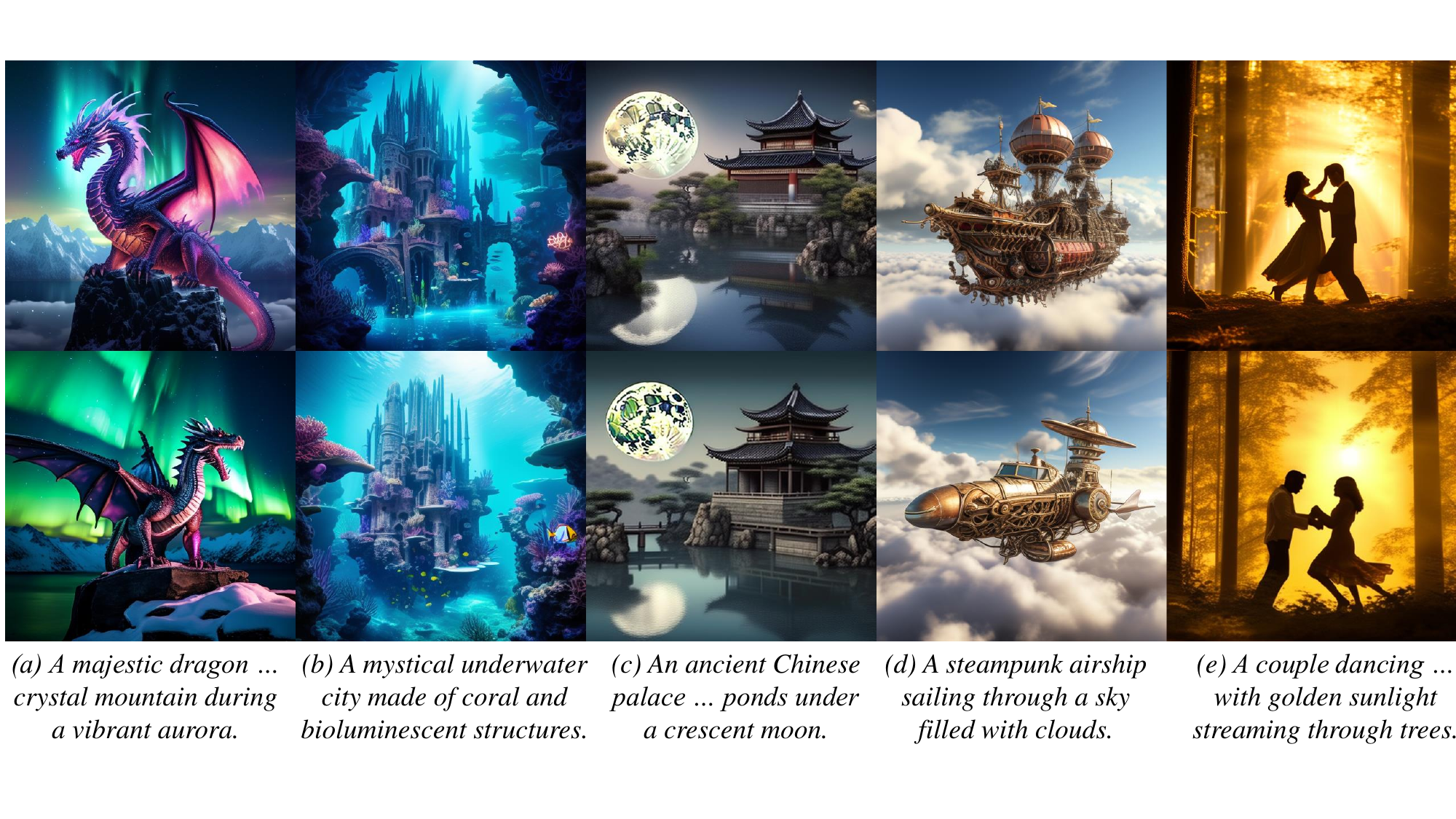}
    \vspace{-10mm}
    \caption{
        {\bf Qualitative results for PixArt-Alpha.} Comparison of T5-XXL (top) and T5-Base (bottom). 
        \label{fig:pixart_comparison}
        }
\end{figure*}

\begin{table*}[!t]
    \centering
    \small 
    \resizebox{\linewidth}{!}{
    \begin{tabular} {l c c c c c c }
    \toprule
    \multicolumn{1}{c}{\multirow{2}{*}{\textbf{Model}}}   
      & \multicolumn{3}{c}{\textbf{Attribute Binding}}     
      & \multicolumn{2}{c}{\textbf{Spatial Relationship}} 
      & \multicolumn{1}{c}{\multirow{2}{*}{Numeracy~($\uparrow$)}} \\
                  \cmidrule(lr){2-4} \cmidrule(lr){5-6}
     \multicolumn{1}{c}{}  & Color~($\uparrow$) & Shape~($\uparrow$) & Texture~($\uparrow$) & 2d-spatial~($\uparrow$) & 3d-spatial~($\uparrow$) \\
    \midrule 
    
    Pixart-Alpha w/ T5-Base & 39.89 & 45.00	& 51.58 & 18.75 & 35.59 & 49.75
 \\
    Pixart-Alpha w/ T5-XXL (teacher) & 38.91 & 40.78 &	46.18 &	20.24&	34.63&	50.29 \\
    
    \bottomrule
    \end{tabular}
    
    }
    \caption{
    \label{table:pixart_compbench}
    {\bf Semantic Understanding Comparison for PixArt-Alpha.} We compare our T5-Base and T5-XXL across the six categories in T2I-CompBench \cite{huang2023t2i}.
    }
\end{table*}

\begin{table}[t]
    \centering
    \small 
    \resizebox{\linewidth}{!}{
    \begin{tabular}{l@{\hspace{1em}}c@{\hspace{1em}}c}
    \toprule
    Model  & FID~($\downarrow$) & CLIP-Score~($\uparrow$) \\
    \midrule
    Pixart-Alpha w/ T5-Base & 33.58 & 28.22 \\
    Pixart-Alpha w/ T5-XXL (teacher) & 29.59 & 30.75 \\
    \bottomrule
    \end{tabular}
    
    }
    \caption{
    \label{table:pixart_clip_fid}
    {\bf FID/CLIP-Score Comparison for PixArt-Alpha.} We compare our T5-Base and T5-XXL on the full MSCOCO-2014 validation set.
    }
\end{table}

\noindent \textbf{Generalizability of Base Models.}
In addition to Flux, we also evaluated our method on PixArt-Alpha using T5-Base as the student encoder. For training, we employed the LAION-Aesthetics-6.5+ and T2I-CompBench datasets, excluding CommonText due to PixArt-Alpha's difficulty in rendering text. We use the 512 checkpoint \cite{pixart512}. The training was conducted on 8 A100 GPUs with a total batch size of 32. The guidance scale was randomly sampled between 5 and 10, in intervals of 0.5, and we used 20 steps for iterative denoising. We applied the AdamW \cite{loshchilov2017decoupled} optimizer with default PyTorch parameters, along with a linear learning rate scheduler and a learning rate of $1e-4$. 

The results, summarized in Tables \ref{table:pixart_compbench} and \ref{table:pixart_clip_fid}, show that our T5-Base model surpasses T5-XXL in categories related to semantic understanding, such as color, shape, texture, and 3D spatial, with only a minimal decrease in performance for 2D spatial and numeracy tasks. In terms of image quality, while T5-Base has a lower CLIP score and a higher FID score compared to T5-XXL, it still demonstrates the ability to generate high-quality images, as depicted in Figure \ref{fig:pixart_comparison}.

\section{More Qualitative Results}
\label{sup:qualitative}
\begin{figure*}[!t]
    \centering
    \includegraphics[width=1\textwidth]{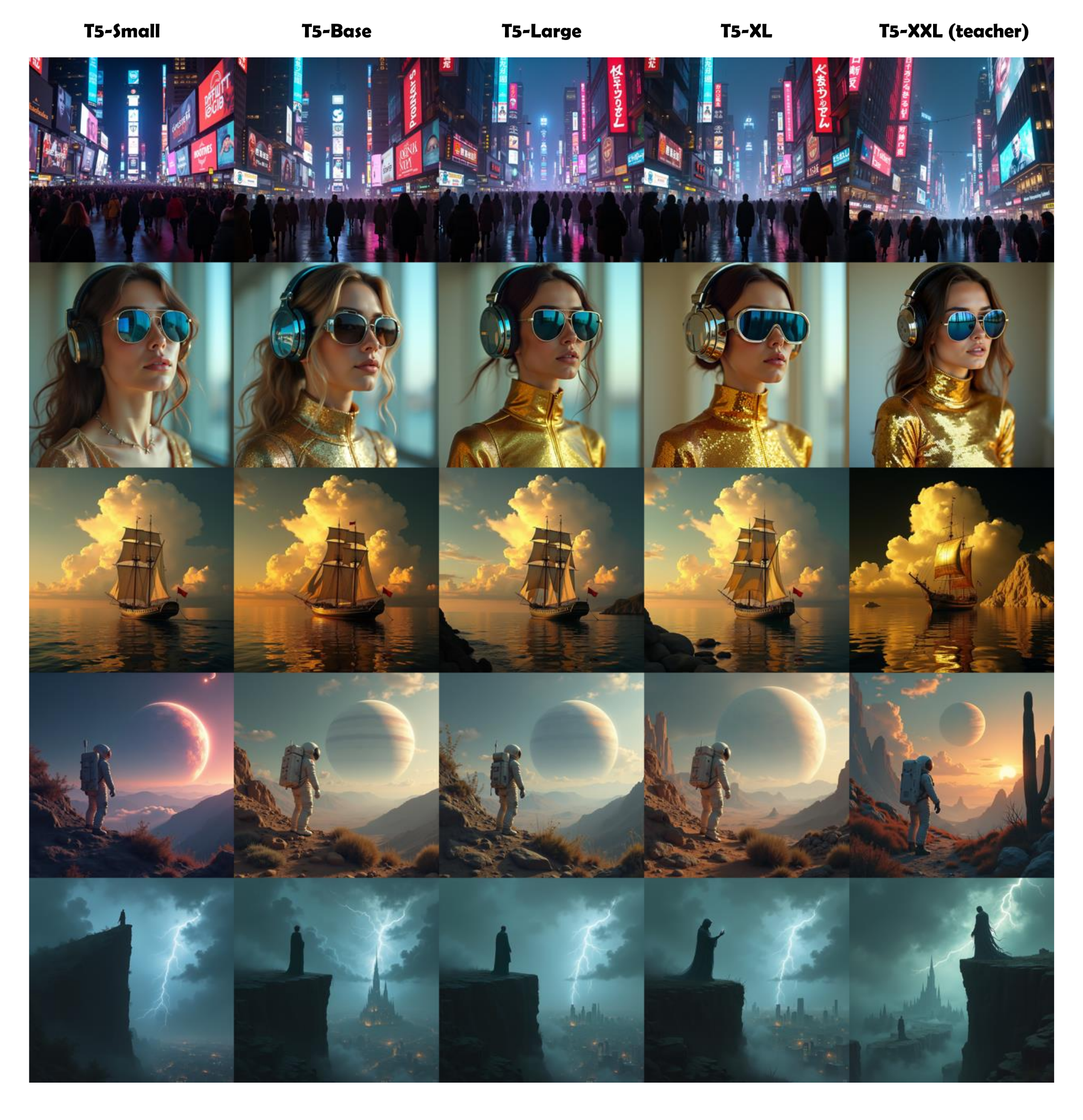}
    \caption{
        {\bf Image Quality Comparison across T5 of Different Size.} We use the same seed and guidance scale of 3.5. (row 1): A bustling cyberpunk metropolis at night, illuminated by a kaleidoscope of neon lights and holographic advertisements. The streets are crowded with people wearing futuristic attire. (row 2): Portrait of a stylish young woman wearing a futuristic golden bodysuit that creates a metallic, mirror-like effect. She is wearing large, reflective blue-tinted aviator sunglasses. Over her head, she wears headphones with metallic accents. (row 3): Baroque ship, beautiful golden mirror sail, golden cumulus clouds, black sky, golden rock, surreal, vivid colors, chiaroscuro lighting, 50mm lens. (row 4): An astronaut exploring a mysterious alien landscape, with strange vegetation and a planet rising in the sky. (row 5): Haunting dark fantasy illustration of an ancient, twisted statue standing atop a steep cliff, overseeing a decaying metropolis shrouded in mist. The sky churns with ominous clouds and flashes of lightning.
        \label{fig:image_quality_comparison}
        }
        \vspace{-3mm}
\end{figure*}

\begin{figure*}[!t]
    \centering
    \includegraphics[width=1\textwidth]{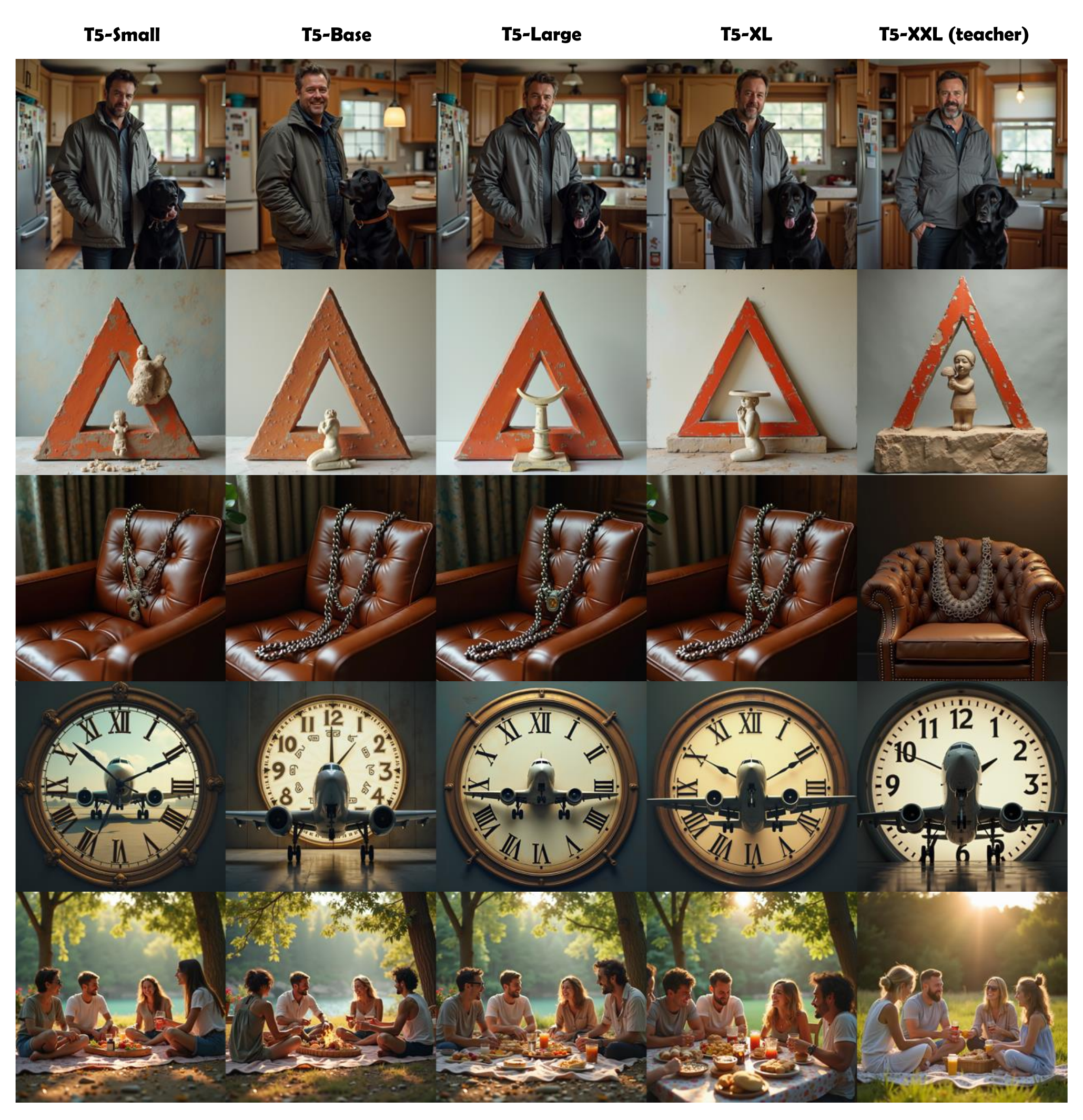}
    \caption{
        {\bf Semantic Understanding Comparison.} Each prompt corresponds to a category of T2I-CompBench, specifically color, shape, texture, 3D-relationship, and numeracy. (row 1): A man in a gray jacket standing in a kitchen next to a black dog. (row 2): A triangular sign and a small sculpture. (row 3): A metallic necklace and a leather chair. (row 4): An airplane in front of a clock. (row 5): Four people gathered for a picnic.
        \label{fig:semantic_comparison}
        }
        \vspace{-3mm}
\end{figure*}

\begin{figure*}[!t]
    \centering
    \includegraphics[width=1\textwidth]{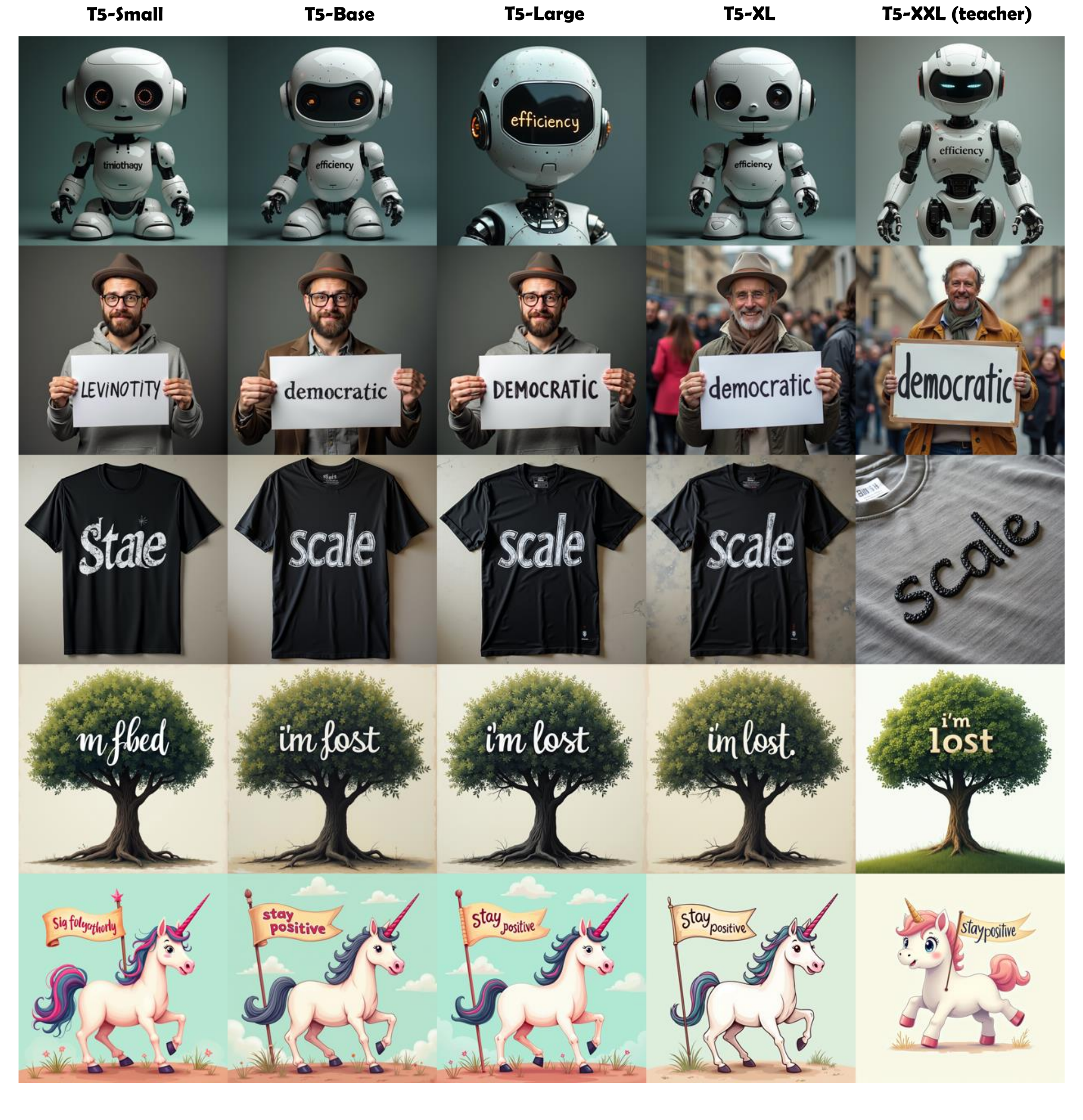}
    \caption{
        {\bf Text Rendering Comparison.} Models larger than T5-Small can render text on various objects within diverse contexts. (row 1): A robot displaying the word 'efficiency'. (row 2): An optimist holding a sign reading 'democratic'. (row 3): A t-shirt with the inscription 'scale'. (row 4): A tree displaying the word 'i'm lost'. (row 5): A unicorn carrying a banner with words 'stay positive'.
        \label{fig:text_rendering_comparison}
        }
        \vspace{-3mm}
\end{figure*}

We provide more qualitative comparison between T5 of different size in \cref{fig:image_quality_comparison,fig:semantic_comparison,fig:text_rendering_comparison}. While T5-Small is capable of generating images with reasonable quality, it occasionally fails to capture the precise semantics of prompts, such as row 2 of Figure \ref{fig:image_quality_comparison} and row 4 of Figure \ref{fig:semantic_comparison}. It completely fails in text rendering as shown in Figure \ref{fig:text_rendering_comparison}. Models larger than T5-Small generally performs well in all three categories, indicating that smaller text encoders such as T5-Base suffice for general image synthesis. However, T5-XXL still excels in generating fine details. When computational resources are not a constraint, T5-XXL is still the preferred text encoder for diffusion models.

\section{Limitations \& Social Impact}
\label{sup:limitation}
\noindent\textbf{Limitations.}
Although our method reduces the memory requirement of GPUs to run large diffusion models, they may not capture the full depth and complexity of larger datasets, potentially leading to a loss in the richness and accuracy of generated content, especially for complex tasks. The process of model compression and knowledge distillation can also introduce or amplify existing biases, as the distilled model might overfit to specific training data characteristics \cite{huang2021sparse,binici2022preventing}. Additionally, the training time is relatively long due to the iterative nature of our step-following distillation. 
A potential solution is to use real images for pre-training. For future work, since LLM are becoming increasing popular for image synthesis \cite{xie2024sana,liu2024playground}, it is worth investigating how we can distill LLM for image generation.

\noindent\textbf{Social Impact.} The development of smaller, efficient text encoders democratizes access to advanced diffusion models, fostering innovation in image synthesis by reducing computational barriers. However, these smaller models might lose some depth and accuracy, potentially introducing or exacerbating biases if not carefully managed. Additionally, increased accessibility might lead to over-reliance on AI-generated content, impacting creativity and originality. Balancing these benefits and drawbacks is crucial for maximizing positive social impact.

\clearpage

\clearpage
\section{Prompts for Generating Images}
\label{sup:prompts}
\subsection{Figure 5 in Main Text}
\begin{enumerate}
    \item A stunning Japanese-inspired fantasy painting of a lone samurai, silhouetted against a massive full moon, standing beneath a windswept, crimson-leafed tree. Falling petals swirl around him, creating a melancholic yet serene atmosphere. The dramatic chiaroscuro lighting highlights the dramatic contrast between the cool-toned background of deep blues and grays and the warm reds of the foliage.
    \item Ink-splash-style. Extreme closeup of a dapper figure in a stylized, richly detailed black top hat, adorned with decorative golden accents, stands against a white background. The character is a skeleton with very detailed skull and long canines as vampire fangs. He cloaked in a vibrant victorian jacket, featuring intricate golden embellishments and a deep red vest underneath. He wears a large victorian monocle with a yellow-tinted lense and copper frame very reddish. Exquisite details include a shiny silver cross and a blue gem on the chest, harmonizing with splashes of paint in vivid hues of blue, gold, and red that artistically cascade around the figure, blending an impressionistic flair with elements of surrealism. The atmosphere is whimsical and opulent, evoking a sense of grandeur and mystery.

    \item An ancient, overgrown temple in a dense jungle, illuminated by the soft light of early morning.
    
    \item A beautiful woman stands gracefully beside her companion, a majestic lion. The lion stands tall and proud, its mane a cascade of alternating black and white gears and intricate cogs, its body radiating a soft, ethereal glow. The woman, radiating grace and elegance, wears a flowing gown of swirling black and white that blends seamlessly with the ethereal landscape. The scene is set against a backdrop of a timeless realm, bathed in the soft glow of a twilight where black and white blend seamlessly.  
    
    \item A portrait of a cybernetic geisha, her face a mesmerizing blend of porcelain skin and iridescent circuitry. Her elaborate headdress is adorned with bioluminescent flowers and delicate, glowing wires. Her kimono, a masterpiece of futuristic design, shimmers with holographic patterns that shift and change, revealing glimpses of the complex machinery beneath. Her eyes gaze directly at the viewer with an enigmatic expression.

    \item A single, crazy blue and black fighter in the sky. It overwhelms the viewer with its artistic flying skills while trailing a meteor tail. Ace pilot of the Republic who was unrivalled in the 1940s. His second name is: The Magician of the Blue Wings, a genius aviator, one of a kind in 100 years. The warriors who challenged him, were destroyed by him, were overrun by him and scattered became many stars. The Milky Way is said to be the graveyard of such aerialists. ‘As we drive our dreams, we fly across the sky and weave our dreams for tomorrow's night.’ He told me with few words. 'The fighter who wants peace more than anyone else, who gives up everything, who flies faster than anyone else. Like a song spinning in the night sky, it pioneers the starry skies, scattering fantastic sparkles.

    \item An incredibly realistic scene of a white kitten wearing a majestic golden veil, rendered with high attention to detail. The cat’s eyes, large and amber, should be given a reflective quality that captures the viewer’s gaze. The golden veil should be richly detailed with intricate patterns, cascading elegantly over the cat’s head and shoulders, with delicate folds that suggest a soft, luxurious fabric. The veil should be adorned with a diadem featuring a prominent red gemstone, surrounded by golden filigree.
    
    \item A high fantasy castle floating among the clouds at sunset, surrounded by flying mythical creatures.

    \item A grand library filled with ancient books and magical artifacts, lit by the warm glow of candlelight.
\end{enumerate}

\subsection{Figure 8.a in Main Text}
\subsubsection{Canny}
\begin{enumerate}
    \item A cyberpunk living room.
    \item A living room with British royal style
    \item A living room in jungle.
    \item A living room with moonlight shedding in.
\end{enumerate}

\subsubsection{Depth}
\begin{enumerate}
    \item A white rabbit in forest.
    \item A furry rabbit resting on grass of a park.
    \item A robot rabbit in a futuristic lab.
    \item A brown rabbit on a table with light shining on its fur.
\end{enumerate}

\subsubsection{Pose}
\begin{enumerate}
    \item A female travel blogger with messy beach waves.
    \item A female adventure photographer with windswept hair.
    \item A male travel influencer with tousled mountain curls.
    \item A business man with suit in a coffee store.
\end{enumerate}

\subsection{Figure 8.b in Main Text}
\begin{enumerate}
    \item Anime ((masterpiece,best quality, detailed)), outdoor, wind lift, souryuu asuka langley, interface headset, red bodysuit, (realistic:1.3).
    
    \item Well-fitting man equipped with hoodie and cap hiding upper face in anime manga style.
    \item Anime theme masterpiece,best quality,1girl,solo,looking at viewer, fur (clothing), black hair, black legwear,(electric guitar:1.4),  reflection, splash, droplets, rust, sparks, asphalt, ground vehicle, sports car, super car, mechanical,burning, playing instrument, livestream.
    \item Anime boy with a dragon companion, standing in a medieval village, with a castle in the background, (heroic:1.3).
    \item Anime girl in a futuristic racing suit, riding a high-tech motorcycle through a neon-lit city, (dynamic:1.3).
    \item An anime girl in a blue dress and straw hat, with long black hair and flowing curly bangs, in the style of anime, against a background of a coastal street by the sea, on a bright sunny day, with flowers on a windowsill, with a cheerful expression, with detailed design, with a watercolor painting effect, and vibrant colors, Hayao Miyazakis manga, with high resolution and clear details --ar 1:2 --stylize 750 --v 6.1.
    \item Pretty cyborg lady, lots of details, sakura flowers, fine art, futuristic setting.
    \item Anime boy with headphones, sitting in a cozy room filled with books and plants, working on a computer, (slice of life:1.3).
    \item Anime warrior princess with a glowing sword, standing in a mystical forest, surrounded by magical creatures, (epic:1.3).
    \item Anime girl with long flowing hair, holding a magical staff, standing on a cliff overlooking a vast ocean, with a sunset in the background, (fantasy:1.3).
\end{enumerate}

\subsection{Figure 8.c in Main Text}
\begin{enumerate}
    \item A fantasy forest with glowing mushrooms and mystical creatures.
    
    \item A bustling market in a medieval village with various stalls and people.
    
    \item A space station orbiting a planet with astronauts floating outside.
    
    \item A dramatic stormy sea with a lighthouse and waves crashing against the rocks.
    
    \item A snowy mountain landscape with a cozy cabin and smoke coming from the chimney.
\end{enumerate}


\end{document}